\documentclass[paper=a4, fontsize=11pt]{scrartcl}
\usepackage[T1]{fontenc}
\usepackage{fourier}

\usepackage[english]{babel}															
\usepackage[protrusion=true,expansion=true]{microtype}	
\usepackage{amsmath,amsfonts,amsthm} 
\usepackage{subcaption}
\usepackage[pdftex]{graphicx}	
\usepackage{url}
\usepackage{lscape}
\usepackage{geometry}
\usepackage[hidelinks]{hyperref}
\usepackage{multirow}
\usepackage{booktabs}
\usepackage{xcolor}

\usepackage{sectsty}
\allsectionsfont{\centering \normalfont\scshape}

\usepackage{fancyhdr}
\numberwithin{equation}{section}		
\numberwithin{figure}{section}			
\numberwithin{table}{section}				

\newcommand{\horrule}[1]{\rule{\linewidth}{#1}} 	

\title{
		\usefont{OT1}{bch}{b}{n}
		\normalfont \normalsize \textsc{Samsung AI Center Montreal \\ LLM-Optimization Team} \\ [25pt]
		\horrule{2pt} \\[0.4cm]
		\huge Hallucination Detection and Hallucination Mitigation: An Investigation \\
		\horrule{2pt} \\[0.5cm]
}
\author{
		\normalfont 	\normalsize
        Junliang Luo,~Tianyu Li,~Di Wu,~Michael Jenkin,~Steve Liu,~Gregory Dudek\\
        \normalfont 	\normalsize
        Samsung AI Center, Montreal, Quebec, Canada\\
        \normalfont 	\normalsize
        junliang.luo@partner.samsung.com,~\{tianyu.li,~di.wu1\}@samsung.com\\
        \normalfont 	\normalsize
        m.jenkin@partner.samsung.com,~\{steve.liu, greg.dudek\}@samsung.com
}
\date{}

\begin{document}
\maketitle

\section*{Abstract}
Large language models (LLMs), including ChatGPT, Bard, and Llama~\cite{touvron2023llama}, have achieved remarkable successes over the last two years in a range of different applications. In spite of these successes, there exist concerns that limit the wide application of LLMs. A key problem is the problem of {\em hallucination\/}. Hallucination refers to the fact that in addition to correct responses, LLMs can also generate seemingly correct but factually incorrect responses. This report aims to present a comprehensive review of the current literature on both hallucination detection and hallucination mitigation. We hope that this report can serve as a good reference for both engineers and researchers who are interested in LLMs and applying them to real world tasks.

\section{Introduction}
Large language models (LLMs) are trained machine learning models that generate symbols (typically text) based on a sequence of previous symbols provided. This previous set of symbols is known as the prompt. The output symbol can then be appended to the prompt and the LLM used to generate the next output symbol. This process is repeated until the complete output is obtained. LLMs have shown significant capabilities, and LLM-based technology can now be found at the core of a number of research projects and commercial AI applications. Although LLMs have shown considerable promise, one concern with their use in critical applications is that the LLMs are prone to ``hallucinations'', that is to produce output that are seemingly correct but contains nonfactual information 

The consequence of hallucinations can be catastrophic and the LLM can produce believable outputs that are misleading and factually incorrect. Detecting and dealing with hallucinations has thus become a critical issue for LLMs to be applied to real-world problems. Here we provide a review of current approaches to detect and mitigate hallucinations. This report is organized as follows. In Section~\ref{sec:intro_metrics}, we present a brief introduction to some common natural language generation (NLG) metrics as well as classification metrics that are presented in these papers. Section~\ref{sec: detection} and Section~\ref{sec: mitigation} present the discussions on existing works on hallucination detection and hallucination mitigation. Finally, we present reproduced results of existing works in Section~\ref{sec: reproduce}. Here we first present a quick overview of the key characteristics of the reviewed papers in Table~\ref{tab:lr_overview} and Table~\ref{tab:summary_mitigation_papers} for hallucination detection and mitigation tasks, respectively. 

\begin{table}[h]
\caption{Summary of the reviewed literature on hallucination detection.}
\resizebox{1\textwidth}{!}{%
\begin{tabular}{l|l|l|l|l|l|l|l}
\hline
Method                                                            & Affiliation & Conference & Code Available                                     & Dataset & Task                                                                 & Key Words                                                                          & Type     \\ \hline
Hades~\cite{liu2022token}                   & Microsoft   & ACL'22     & \href{https://github.com/microsoft/HaDes}{\textcolor{blue}{Yes}}                         & \href{https://github.com/microsoft/HaDes}{\textcolor{blue}{Yes}}     & QA                                                                   &  \begin{tabular}[c]{@{}l@{}}LLM Fine-tune, synthetic dataset,\\ contextual detector\end{tabular}                               & Token    \\ \hline
NPH~\cite{dziri2021neural}                  & U Alberta   & EMNLP'21   & \href{https://github.com/nouhadziri/Neural-Path-Hunter}{\textcolor{blue}{Yes}}           & No      & QA                                                                   & LLM fine-tune, contextual detector, knowledge grounded                             & Token    \\ \hline
Enfa~\cite{cao-etal-2022-hallucinated}      & MILA        & ACL'22     & \href{https://github.com/mcao516/EntFA}{\textcolor{blue}{Yes}}                          & \href{https://github.com/mcao516/EntFA}{\textcolor{blue}{Yes}}     & Summarization                                                        & Prior and posterior probabilities, k-nn, human annotated dataset                   & Token    \\ \hline
CNSG~\cite{zhou2021detecting}               & FAIR        & ACL'21     & \href{https://github.com/violet-zct/fairseq-detect-hallucination}{\textcolor{blue}{Yes}} & \href{https://github.com/violet-zct/fairseq-detect-hallucination}{\textcolor{blue}{Yes}}     & \begin{tabular}[c]{@{}l@{}}Summarization\\ Translation\end{tabular}  & LLM Fine-tune, synthetic dataset, contextual detector                              & Token    \\ \hline
SelfCheckGPT~\cite{manakul2023selfcheckgpt} & U Cambridge & N/A        & \href{https://github.com/potsawee/selfcheckgpt}{\textcolor{blue}{Yes}}                  & \href{https://github.com/potsawee/selfcheckgpt}{\textcolor{blue}{Yes}}      & QA                                                                   & Sampling-based, self-consistency, human annotated dataset, plug-n-play             & Sentence \\ \hline
AlignScore~\cite{zha-etal-2023-alignscore}  & UCSD        & ACL'23     & \href{https://github.com/yuh-zha/AlignScore}{\textcolor{blue}{Yes}}                      & No      & Various NLG tasks                                                    & Unification of data sources and tasks, LLM fine-tune                               & Sentence \\ \hline
ExHalder~\cite{shen2023misleading}          & Google      & WWW'23     & No                            & \href{https://bit.ly/exhalder-dataset}{\textcolor{blue}{Yes}}     & Summarization                                                        & NLI dataset pretraining, augmented fine-tuning, human annotated dataset            & Sentence \\ \hline
Harim+~\cite{son-etal-2022-harim}           & NCSOFT      & AACL'22    & \href{https://huggingface.co/spaces/NCSOFT/harim_plus}{\textcolor{blue}{Yes}}           & No      & Summarization                                                        & \begin{tabular}[c]{@{}l@{}}Plug-n-play, prior and posterior probabilities,\\ decoder overconfidence regularizer\end{tabular} & Sentence \\ \hline
HaluEval~\cite{HaluEval}                    & Renmin U    & N/A        & \href{https://github.com/RUCAIBox/HaluEval}{\textcolor{blue}{Yes}}                      & \href{https://github.com/RUCAIBox/HaluEval}{\textcolor{blue}{Yes}}     & \begin{tabular}[c]{@{}l@{}}QA, dialog, \\ summarization\end{tabular} & Benchmark datasets, LLM self-detection, prompt engineering                         & Sentence \\ \hline
\end{tabular}%
}
\label{tab:lr_overview}
\end{table}

\begin{table}[h]
\caption{Summary of the literature reviewed on hallucination mitigation.}
\resizebox{1\textwidth}{!}{%
\begin{tabular}{l|l|l|l|l|l|l}
\hline
Method                                                            & Affiliation & Conference & Code Avaiable                                     & Dataset & Task                                                                 & Key Words \\ \hline

RHO~\cite{ji2023rho} & HKUST & ACL '23 & \href{https://github.com/ziweiji/rho}{\textcolor{blue}{Yes}} & No & KGD & Conversational model, local and global grounding, human evaluation \\ \hline

RAGs~\cite{shuster2021retrieval} & Facebook & EMNLP '21 & \href{https://parl.ai/projects/hallucination/}{\textcolor{blue}{Yes}} & No & KGD & Retrieval-Augmented Generation (RAG), encoder-decoder architecture, human evaluation  \\ \hline

Control Codes~\cite{shuster2021retrieval} & Google & ACL 2021 & No & No & KGD & Control codes (Guide tokens), LLM training and output resampling. human evaluation \\ \hline

MixCL~\cite{sun2023contrastive} & U Shandong & AAAI '23 & \href{https://github.com/sunnweiwei/mixcl}{\textcolor{blue}{Yes}} & No & KGD & Contrastive Learning, LLM fine-tune, human evaluation \\ \hline

HERMAN~\cite{zhao2020reducing} & U Edinburgh & EMNLP 2020 & No & No & Summarization & Quantitative correction. Bidirectional LSTM encoder, attention mechanism, human evaluation \\ \hline

Self-contradictory~\cite{mundler2023self} & ETH Zurich & N/A & \href{https://github.com/eth-sri/chatprotect}{\textcolor{blue}{Yes}} & \href{https://github.com/eth-sri/ChatProtect/commit/504f5b0b07cc3eb5cf528752cb6cb9bc6731d68b}{\textcolor{blue}{Yes}} & QA & Self correction, prompt engineering, LM-based analyzer \\ \hline

\end{tabular}%
}
\label{tab:summary_mitigation_papers}
\end{table}

\section{Introduction to Common Metrics}
\label{sec:intro_metrics}
This report is primarily concerned with hallucination detection and mitigation. This section reviews some of the common metrics on classification and more general natural language generation tasks. 

\subsection{Classification Metrics}
Classification problems take an observation and identify which of a set of categories to which it belongs. Under the scope of this report, the classification problem is often presented as a binary classification problem, i.e., either the response is hallucinated or not hallucinated. In this subsection, we briefly introduce  some of the commonly-seen classification metrics. 

\paragraph{Accuracy} The classification accuracy is the ratio of the number of correct predictions to the total number of inputs.

\paragraph{Precision and recall} Precision and recall are a popular pair of metrics in the classification literature. Assume we are under the binary classification setup, i.e., either positive class or negative class. Then precision is essentially asking the question  ``What proportion of positive predictions is actually correct?,'' while recall is asking the question  ``What proportion of actual positives is classified correctly?''. 

\paragraph{F-score} To fully evaluate the effectiveness of a model, one must examine both precision and recall. Unfortunately, precision and recall are often in tension. That is, improving precision typically reduces recall and vice versa. One classic approach to this issue is to use F-score. The F1 score is a specific F score. F1 is the harmonic mean between recall and precision. The harmonic mean is bounded. It  attends more towards the lower value. Therefore, even if one of recall and precision is high, F1 will be more responsive the lower one and won't be driven up by just one high score. 

\paragraph{AUC}
Often the classification model will assign scores for the instances to be evaluated. A final classification result can be obtained by thresholding these scores. As a consequence, different threshold often leads to different classification result. A ROC curve (receiver operating characteristic curve) is a graph showing the performance of a classification model at all classification thresholds. This curve plots two parameters: the true positive rate and the false positive rate, where the former is a synonym for recall, and the later describes the proportion of all instances that are falsely classified as positive with respect to the negative instances. AUC stands for ``Area under the ROC Curve'',  that is, AUC measures the two-dimensional area underneath the entire ROC curve. 

\paragraph{BSS}
The BSS (Brier Skill Score) is a score function that measures the accuracy of probabilistic predictions. For unidimensional predictions, the BSS is strictly equivalent to the mean squared error as applied to predicted probabilities. Given a set of instances and their true class, BSS computes the mean squared difference between the predicted probability assigned to the possible classification result for each instance and the actual class of the instance. 

\paragraph{G-Mean} 
The geometric mean (G-mean) is the root of the product of class-wise sensitivity. This measure tries to maximize the accuracy on each of the classes while keeping them balanced. For binary classification G-mean is the squared root of the product of the sensitivity and specificity. For multi-class problems it is a higher root of the product of sensitivity for each class.

\subsection{NLG Metrics}
\label{sec:metric}
Typically, when evaluating an NLG (natural language generation) system, one is given a candidate text sequence (prediction) and a (list of) reference text (the ground truth) and the similarity of the two should be evaluated. For example, a candidate/reference text pairs could be given by the following:

\begin{itemize}
    \item Candidate: "I am a member of SAIC Montreal."
    \item Reference A: "I am from Montreal."
    \item Reference B: "This is SAIC Montreal."
\end{itemize}
where reference A and B together forms a list of reference text. The goal of the evaluate is to compute the similarity/distance between the candidate and the reference text. 
In this subsection, we will briefly talk about some of the common metrics used in NLG tasks, especially regarding to this report. 

\paragraph{N-Grams} N-grams are a series of adjacent words, symbols or tokens in a given document. When used as an NLG metric, n-grams often refers to the fraction of n-grams in the candidate text which are present in any of the reference texts. For example, assume we are given the candidate/reference pair as in the example shown above, the total number of 2-grams in the reference text is 6, while the overlapped 2-grams are ``I am'' and ``SAIC Montreal'', which gives a 2-grams score of $1/3$. 

\paragraph{BLEU} BLEU~\cite{papineni2002bleu} (Bilingual Evaluation Understudy) is a metric for automatically evaluating various NLG tasks with reference text. The BLEU score is a number between zero and one that measures the similarity of the candidate text to a set of reference texts. A value of 0 means that the candidate output has no overlap with the reference text (low quality) while a value of 1 means there is perfect overlap with the reference (high quality). BLEU is a precision based score, meaning it looks at the fraction of overlapped n-grams between the candidate and the reference w.r.t. the total n-grams of the candidate text. One can see that it is easy to cheat by generating very short candidate text. Therefore, to circumvent this issue, BLEU proposes to incorporate a brevity penalty to penalize short candidate text.

\paragraph{SACREBLEU} Standard BLEU often depends on the tokenization methods used, resulting in non-comparable BLEU scores across different experimental setups. SacreBLEU\cite{post-2018-call} uses standardized tokenization and normalization schemes to unify the metrics to allow cross-experiment comparisons. 

\paragraph{Rouge} Rouge~\cite{lin-2004-rouge} is the abbreviation of Recall Oriented Understudy for Gisting Evaluation. As is clear from its name, ROUGE is based only on recall, and is often used for summary evaluation. The computation is similar to BLEU, except that instead of having the number of n-grams in the candidate as the denominator, Rouge uses the number of n-grams in the references as the denominator. Depending on the feature used for calculating recall, ROUGE may be of many types, namely ROUGE-N, ROUGE-L, ROUGE-W, and ROUGE-S. 

\paragraph{Meteor} METEOR~\cite{banerjee-lavie-2005-meteor} (Metric for Evaluation for Translation with Explicit Ordering) is another metric for machine translation evaluation. One problem with BLEU is that it  is often considered as being too rigid as it only allows exact n-gram matching. To address this problem, METEOR modifies the precision and recall computations, replacing them with a weighted F-score based on mapping unigrams with relaxed measure on synonym and paraphrase matching as well as stemmed-word matching. It also adds a penalty function for incorrect word order.

\paragraph{BERTScore} The previously described metrics are word-overlap-based metrics. That is, they are computed via the percentage of the overlapped text spans between the candidate and the references. Another popular type of metrics is the embedding based metrics. Word embeddings extend word overlap-based metrics beyond exact match. Instead of just looking at the frequencies of overlapped words, we now look at the distance (inner product or cosine similarity) between the embeddings of these words. BERTScore~\cite{bert-score} is an embedding-based metric. BERTScore computes the similarity of two sentences as a sum of cosine similarities between their tokens' embeddings using a BERT model. 

\paragraph{BARTScore} BARTScore~\cite{NEURIPS2021_bartscore} conceptualizes the evaluation of generated text as a text generation problem, modeled using pre-trained sequence-to-sequence models~(BART~\cite{lewis2020bart}). The core idea is that models trained to convert the generated candidate text to/from a reference output will achieve higher scores when the candidate text is better. BARTScore is parameter- and data-efficient. It can also better support evaluation of generated text from different perspectives (e.g., informativeness, coherence, factuality, etc.).

\paragraph{BLEURT} 
BLEURT~\cite{sellam2020bleurt, pu2021learning} (Bilingual Evaluation Understudy with Representations from Transformers) is a machine learning-based automatic metric that can capture non-trivial semantic similarities between sentences. BLEURT leverages the word representations from BERT~\cite{kenton2019bert}, and transfer learning to construct this automatic metric. They first create a large set of synthetic datasets by perturbing the text from Wikipedia and then use automatic metrics, such as BLEU and Rouge to score these texts. Then they construct a regression tasks on these synthetic data with BERT representations. Finally, they finetuned the regression model with human annotations. 

\paragraph{FEQA} 
Similar to BLEURT, FEQA~\cite{durmus-etal-2020-feqa} (Faithfulness Evaluation with Question Answering) is a model-based metric, leveraging the Question-Answering framework. The main idea of this metric is to generate questions and answers from the candidate text, then conditioned on the reference to obtain the answers for the same set of questions and see if the answers match. The advantage of this method is that it aligns really well with human judges on faithfulness. The disadvantage is however, since it is based on LLM~(BART~\cite{lewis2020bart} and BERT~\cite{kenton2019bert}), therefore requires quite significant computing resources, and often slow to run. 

\paragraph{QuestEval}
QuestEval~\cite{scialom-etal-2021-questeval} is also a question-answering based metric, similar to FEQA. One thing to note is that FEQA is a precision-based metric, as it extract questions and answers from the candidate and then ask the reference the same questions to see if their answers match. QuestEval additionally incorporates recall into the framework as well.  For the precision, similar to FEQA, one can generate question and answers from the candidate and then ask the reference for answers and compare to obtain precision score. For the recall, one can generate question from the reference text and conditioned on the candidate to obtain the corresponding answers. Then the QuestEval score can be obtained by combining the precision and recall scores. 

\paragraph{DAE} DAE~\cite{goyal-durrett-2020-evaluating} (Dependency Arc Entailment) is another model-based metric. Unlike FEQA and BLEURT, it leverages the framework of textual entailment. Under this framework, a model is often used to evaluate if the candidate is entailed by the reference. Most of the methods that fall under this category often considers a sentence-level evaluation, however this type of approach do not localize and failed to provide fine-grained evaluation. This work instead, asks whether the semantic relationship manifested by individual dependency arcs in the candidate text is supported by the reference. This approach views dependency arcs as semantic units that can be interpreted in isolation. Each arc is therefore judged independently based on whether the relation it implies is entailed by the source sentence. 

\section{Hallucination Detection}
\label{sec: detection}
Hallucination detection is the task of identifying potential hallucinations in the generated responses by LLMs. Typically in the literature, there are two major archetypes of this task: token-level detection and sentence-level detection. The token-level tasks try to identify the exact token/named entity that might suffer from hallucination while the sentence-level tasks try to identify the hallucinated sentences. In this section, we will review papers of these two types of hallucination detection tasks. 
\subsection{Token Level}
\subsubsection[Hades: A token-level reference-free hallucination detection benchmark for free-form text generation]{Hades: A token-level reference-free hallucination detection benchmark for free-form text generation~\cite{liu2022token}}
\textbf{Github Link: } https://github.com/microsoft/HaDes
\newline
\textbf{Tasks: } Question answering.
\newline
\textbf{Core Idea: } Propose a novel token-level reference-free hallucination detection annotated dataset. Additionally, a series of token-level detection models baselines are implemented and evaluated on the proposed dataset. 
\newline
\newline 
\begin{table}[t]
\centering
\caption{Benchmark (numbers in percentages ($\%$)) for the online setting on HADES, where detecting models have access to the bidirectional context. Except for BSS, all are the higher the better.~(Table source:~\cite{liu2022token})}
\label{tab:hades_online}
\resizebox{\textwidth}{!}{%
\begin{tabular}{lcccccccccc}
\hline
\multirow{2}{*}{Model} & \multirow{2}{*}{ACC} & \multirow{2}{*}{G-Mean} & \multirow{2}{*}{BSS} & \multirow{2}{*}{AUC} & \multicolumn{3}{c}{Not Hallucination} & \multicolumn{3}{c}{Hallucination} \\
 &  &  &  &  & P & R & F1 & P & R & F1 \\ \hline
GPT-2 & \textbf{71.58} & \textbf{70.98} & 19.13 & 77.71 & \textbf{71.32} & 77.29 & \textbf{74.19} & \textbf{71.93} & \textbf{65.19} & \textbf{68.40} \\
BERT & 71.00 & 70.43 & \textbf{18.66} & \textbf{78.83} & 70.91 & 76.50 & 73.60 & 71.12 & 64.84 & 67.84 \\
RoBERTa & 70.67 & 70.14 & 19.77 & 77.07 & 70.74 & 75.87 & 73.22 & 70.58 & 64.84 & 67.59 \\
XLNet & 70.08 & 69.17 & 19.76 & 76.59 & 69.39 & \textbf{77.60} & 73.27 & 71.08 & 61.66 & 66.04 \\ \hline
\end{tabular}%
}
\end{table}

\begin{table}[t]
\centering
\caption{Benchmark (numbers in percentages ($\%$)) for the offline setting on HADES, where detecting models have access to the bidirectional context. Except for BSS, all are the higher the better.~(Table source:~\cite{liu2022token})}
\label{tab:hades_offline}
\resizebox{\textwidth}{!}{%
\begin{tabular}{lcccccccccc}
\hline
\multirow{2}{*}{Model} & \multirow{2}{*}{ACC} & \multirow{2}{*}{G-Mean} & \multirow{2}{*}{BSS} & \multirow{2}{*}{AUC} & \multicolumn{3}{c}{Not Hallucination} & \multicolumn{3}{c}{Hallucination} \\
 &  &  &  &  & P & R & F1 & P & R & F1 \\ \hline
LR & 62.25 & 60.77 & - & - & 62.35 & 72.08 & 66.86 & 62.10 & 51.24 & 60.33 \\
SVM & 63.67 & 61.50 & - & - & 62.89 & 76.18 & 68.90 & 65.05 & 49.65 & 56.31 \\
BERT & 71.92 & \textbf{71.95} & 19.06 & 78.63 & \textbf{74.46} & 71.29 & 72.84 & 69.31 & \textbf{72.61} & 70.92 \\
RoBERTa & \multicolumn{1}{l}{\textbf{72.83}} & \multicolumn{1}{l}{70.94} & \multicolumn{1}{l}{\textbf{18.78}} & \multicolumn{1}{l}{78.72} & \multicolumn{1}{l}{74.06} & \multicolumn{1}{l}{74.46} & \multicolumn{1}{l}{74.41} & \multicolumn{1}{l}{71.43} & \multicolumn{1}{l}{70.67} & \multicolumn{1}{l}{\textbf{71.05}} \\
XLNet & 72.33 & 71.39 & 18.79 & \textbf{78.93} & 71.15 & \textbf{80.13} & \textbf{75.37} & \textbf{74.07} & 63.60 & 68.44 \\ \hline
\end{tabular}%
}
\end{table}
This work centers on detecting hallucinations at the token-level without relying on ground-truth references. While existing methods often operate at the sentence or document level with an oracle reference, these approaches face challenges when ground-truth references are unavailable for free-form text generation. In addition, detecting at finer token levels can be crucial for real-time prevention of misleading content as well.  This work provides two important initial steps towards hallucination detection under this setup. First, the authors develop a novel token-level reference-free hallucination detection annotated dataset named HADES (HAllucination DEtection dataSet). They then create a series of baseline token-level detection models trained on this dataset. These models can detect potential hallucinations at the token level without the need for an oracle reference. This approach could be useful in real-time applications where ground-truth references may not be readily available.

To construct the HADES dataset, roughly three steps are followed. First, a large number of text segments extracted from English language Wikipedia (English WIKI-40B~\cite{guo2020wiki}) using BERT~\cite{kenton2019bert} are perturbed. In applying this contextual perturbation the authors maintain two principles: 1) the fluency and syntactic correctness of the perturbed text should be preserved; 2) the perturbed text should be lexically diverse. The text perturbation process is split into three operations, namely MASK, REPLACE and RANK. In the MASK step,  one word (named entity) spans are masked based on a predefined mask ratio. Then, in the REPLACE step, a pretrained BERT-base model is leveraged to predict the masked span and replace the span with the prediction. All top-k (k=10) samplings are considered as possible candidates as this provides a good trade-off between diversity (number of distinct tokens) and coherence (number of incoherent perturbations). In the final step,  the perturbed text is post-processed with a RANK operation as an additional screening step to filter out global incoherence and syntactic issues. In the end, the authors collect around 1M perturbed text segments in the pool after contextual perturbation. One thing to note is that not all of these contain hallucination, as the BERT model can generate factual information given that it is pretrained on a rich open web corpus.

After the initial data perturbation, the authors use human annotation for the labeling process. Here, they use a so-called ``model-in-the-loop'' procedure to annotate a challenging subset, as human annotation is prohibitively expensive at this scale. The annotation process is split into several rounds. For each round, they first retrain a hallucination detection model (initialized with BERT) on the annotated instances from the previous rounds. This model is used to select the next batch of data to be annotated from the remaining unlabeled data. In total, after accumulating annotations for several rounds, they obtain 12,719 instances with 71,226 HITS from judges. They conduct 14 rounds of annotation, increasing the annotation scale with each round (ranging from around 200 instances/round to around 4000 instances/round). Out of 12,719 annotated instances, 10,954 instances reached consensus among judges and are included in the HADES dataset. The dataset is split into train, validation and test sets with sizes of 8754, 1000, 1200 respectively. In the final dataset, ``hallucination'' cases slightly outnumber ``not hallucination'' cases, with a ratio of $54.5\%/45.5\%$.


Methodology-wise, this work proposes a series of baseline methods based on pretrained transformer models, including BERT~\cite{kenton2019bert}, GPT-2~\cite{radford2019language}, XLNet~\cite{yang2019xlnet} and RoBERTa~\cite{zhuang-etal-2021-robustly}. These transformer-based models can potentially leverage context or embedded world knowledge to detect self-contradictory or anti-commonsense content. Specifically, for an input text segment and a target token, the model predicts binary hallucination labels for the given text spans. The authors propose two different settings: online and offline. In the online setting, the model can only access the unidirectional preceding context, which simulates on-the-fly generation. While in the offline setting, it is assumed that generation is complete, so the the model is able to perceive the bidirectional context. The results are shown in Figure~\ref{tab:hades_online} and Figure~\ref{tab:hades_offline} for online and offline settings respectively. 

One thing worth mentioning is that for the baseline transformer-based detectors, longer context length tends to result in a better performance. The authors run BERT-large detection model with different context lengths and characterize its performance in both online and offline settings. Starting from the target words, they set a fixed size (5/10/20/40/80/160) context window and truncate all text beyond this window. As the context window enlarges, the model performance grows rapidly when context length is smaller than 80, and then gradually converges. This observation highlights the importance of context in hallucination detection using the detectors in this work. Therefore, one should pay attention to the use case of the HADEs detection model that it might underperform when the text length is not sufficiently long.

\subsubsection[NPH: Neural Path Hunter: Reducing Hallucination in Dialogue Systems via Path Grounding]{NPH: Neural Path Hunter: Reducing Hallucination in Dialogue Systems via Path Grounding~\cite{dziri2021neural}}
\textbf{Github Link: } https://github.com/nouhadziri/Neural-Path-Hunter
\newline
\textbf{Tasks: } Question and answering.
\newline
\newline
See Section~\ref{sec:nph}. 


\begin{figure}[t]
        \centering
        \includegraphics[width=0.7\linewidth]{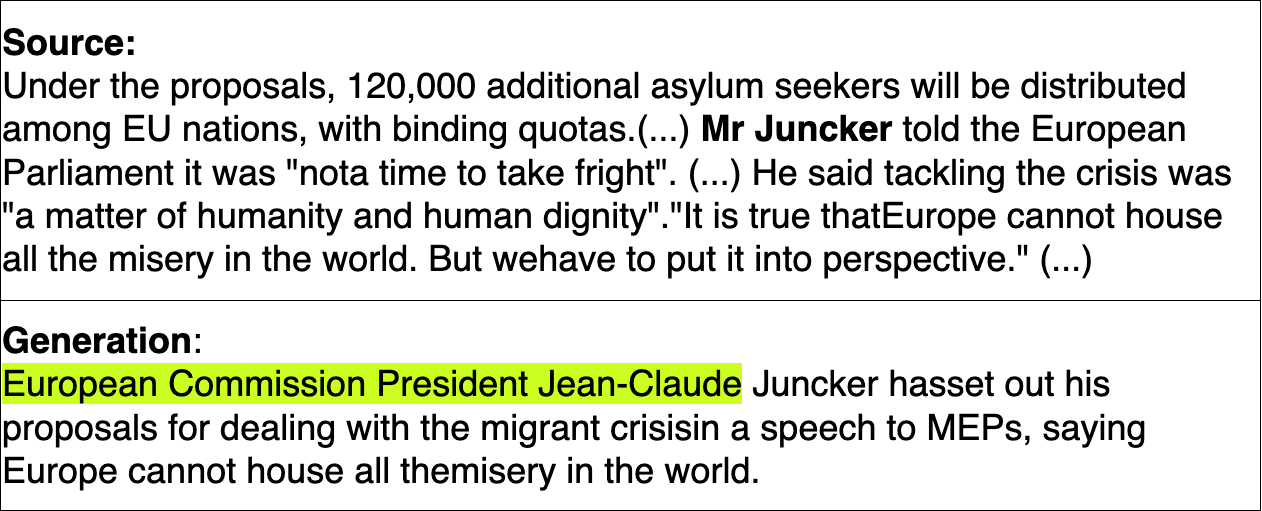}
        \caption{Example of factual hallucinations in a BART generated summary on XSum dataset~\cite{narayan-etal-2018-dont}. Neither the title "European Commission President" nor the first name "Jean- Claude" is mentioned in the document but both are factual.~(Figure source:~\cite{dziri2021neural})}
        \label{fig:factual_hal}
\end{figure}

\subsubsection[Hallucinated but Factual! Inspecting the Factuality of Hallucinations in Abstractive Summarization]{Hallucinated but Factual! Inspecting the Factuality of Hallucinations in Abstractive Summarization~\cite{cao-etal-2022-hallucinated}}
\label{sec:enfa}
\textbf{Github Link: } https://github.com/mcao516/EntFA
\newline
\textbf{Tasks: } Summarization.
\newline
\textbf{Core Idea: } The paper proposes that not all hallucinations are detrimental within the scope of text summarization. On the contrary, some hallucinations can be factual and can enrich the semantics of the summarized text. 
\newline
\newline
Previous methods (Hades, NPH) mainly address the hallucination detection problem under the scope of the question and answering task. Under this setting, hallucination is often defined to be answers that are factually false. In summarization, however, the definition is slightly different as the summarization task involves extracting and abstracting key content from the source document. Therefore, hallucination under this task is often defined to be any content in the summary that is not directly inferrable from the source text. One of the main contributions of~\cite{cao-etal-2022-hallucinated} is that the authors point out that some hallucinations under summarization tasks can be factual as well, and these hallucinations should not be removed or addressed in the summaries as they provide additional information that the source text does not contain. One illustrative example can be found in Figure~\ref{fig:factual_hal}. In addition, this work also proposes a novel detection method that separates factual from non-factual hallucinations of entities. This approach utilizes an entity's prior and posterior probabilities according to pretrained and finetuned masked language models, respectively. They show that empirically this method outperforms five baseline methods and strongly correlates with human judges. The authors also provide a small annotated dataset for token-level hallucination detection on summarization task. A Token-level Reference-free Hallucination Detection Benchmark for Free-form text generation.
\begin{table}[t]
    \centering
    \caption{Summary-level Pearson correlation coefficients between various automatic metrics and human judgments of factuality for XSUM datasets. High Pearson correlation coefficients indicate better alignment with human judges.~(Table source:~\cite{dziri2021neural})}
    \label{tab:enfa_results}
    \resizebox{0.5\textwidth}{!}{%
    \begin{tabular}{l|l|l}
    \hline
    Metric & \begin{tabular}[c]{@{}l@{}}FRANK\\ (Partial Pearson's $\rho$)\end{tabular} & PCC \\ \hline
    BLUE & 0.139 & 0.118 \\
    ROUGE-1 & 0.155 & 0.132 \\
    BERTScore & -0.0359 & 0.025 \\
    QAGS & -0.0225 & 0.175 \\
    FEQA & 0.0242 & - \\
    DAE & 0.0444 & - \\ \hline
    ENFA (proposed) & \textbf{0.183} & \textbf{0.268} \\ \hline
    \end{tabular}%
    }
    \end{table}

For the dataset, the authors annotated 800 summaries generated by BART, which was one of the state-of-the-art abstractive summarization models available at the time. The input documents are randomly selected from XSUM test set~\cite{narayan-etal-2018-dont}. 2,838 entities are extracted from the 800 generated summaries. Then $30\%$ of the samples is set aside as the test set, while remaining is for training. The authors manually labeled each entity with one of the following three tags: non-hallucinated, factual hallucination, and non-factual hallucination. First, they extract entities from the given summary using automatic NER tools~\cite{spacy2}. Then extracted entities are labeled based on the following criteria: if the the entity can be directly determined by the information from the source document, then this entity is non-hallucinated; if no internet search can prove or disprove the factuality of the entity or one can find evidence that disprove the factuality of the entity, it is labeled as non-factual hallucination; otherwise, the entity is labeled as factual hallucination. They name this dataset as XENT.

The main task involves differentiating non-factual hallucinations from both factual hallucinations and non-hallucinations. To do so, the authors propose to compute the prior and posterior probabilities of the entity, which is calculated by masked language models based on BART~\cite{lewis2020bart}. The prior probability is defined as the probability of its generation by a language model that does not have access to the source text, while the posterior probability is computed by conditioning on the source document. Intuitively, if the two probabilities are close to one another, then there is a low likelihood that the entity is a factual error (non-factual hallucination) as the small difference indicates providing the source text does not offer more (less) evidence for the entity. To classify the hallucination and factuality statuses of a given entity, the authors propose to use  K-Nearest Neighbors (KNN) as a discriminator model. The reason for using this algorithm, claimed by the authors, is that it requires no training and makes minimal assumptions about the form of the decision boundary, as a non-parametric method and it also offers adequate interpretability. The KNN classifier is trained using the prior and posterior probabilities as features on the labeled dataset. One can find the corresponding experimental results using the XENT dataset can be found in Figure~\ref{tab:enfa_results}. The results show that the proposed approach outperforms five baselines on the factuality classification task.

\begin{figure}
    \centering
    \includegraphics[width=1.0\linewidth]{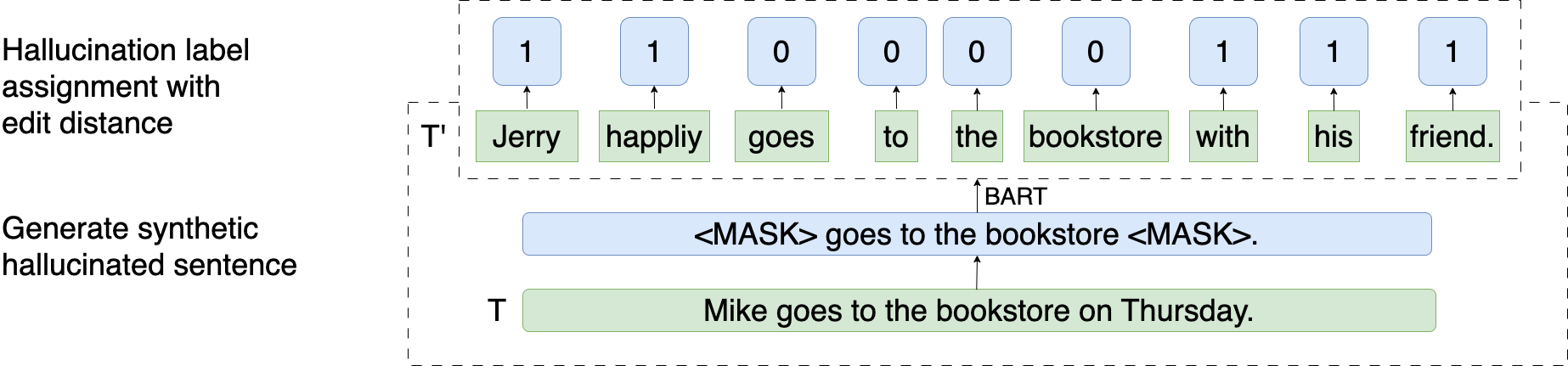}
    \caption{Generation of synthetic data with hallucination labels. A hallucinated version of the original text is generated by feeding the noised sentence to the encoder-decoder model BART. Hallucination labels are assigned to each token by computing the edit distance between the hallucinated text and the original one. Labels of $1$ refer to hallucinated words.~(Figure source:~\cite{dziri2021neural})}
    \label{fig:zhou_dataset}
\end{figure}

\subsubsection[Detecting Hallucinated Content in Conditional Neural Sequence Generation]{Detecting Hallucinated Content in Conditional Neural Sequence Generation~\cite{zhou2021detecting}}
\textbf{Github Link: } https://github.com/violet-zct/fairseq-detect-hallucination
\newline
\textbf{Tasks: } Summarization and translation.
\newline
\textbf{Core Idea: } The paper proposes a pipeline to create token-level hallucinated synthetic dataset. In addition, it also proposes a detection method based on a pretrained language model.
\newline
\newline
In this study, the authors present a technique to identify hallucinations. The method involves fine-tuning pretrained language models on synthetic data, incorporating automatically inserted hallucinations. Through experiments conducted on machine translation (MT) and abstractive summarization, the proposed approach consistently surpasses robust baselines across all benchmark datasets. Additionally, the authors demonstrate the utility of token-level hallucination labels in defining a nuanced loss function for the target sequence in low-resource MT, leading to noteworthy enhancements compared to strong baseline methods. The method is also applied to word-level quality estimation for MT, showcasing its effectiveness in both supervised and unsupervised settings.

Figure~\ref{fig:zhou_dataset} shows the synthetic dataset creation procedure proposed in this paper. This process can be divided into two phases: generation of hallucinated sentences and label assignments. To generate hallucinated sentences, the authors propose to first apply a noising function that removes words from the original target sentence and then use a pretrained BART~\cite{lewis2020bart} to generate a hallucinated version of the original text conditioned on the masked text with beam search. Then, to assign the labels, the authors compute the edit distance between the hallucinated text and the original one, and back-trace the deletion and substitution operations with dynamic programming. All the positions in the hallucinated text involving these two operations are labeled as hallucinations and everything else is considered faithful to the original text. For the abstractive summarization task, the authors perform this procedure on the XSUM dataset~\cite{maynez2020faithfulness}, which comprises 226,711 British Broadcasting Corporation (BBC) articles paired with their single-sentence summaries. For the MT task, they leverage the multi-domain Chinese-English (Zh-En) translation dataset~\cite{wang2020go}, which consists of four balanced domains: law, news, patent and subtitles.

\begin{figure}
    \centering
    \includegraphics[width=1.0\linewidth]{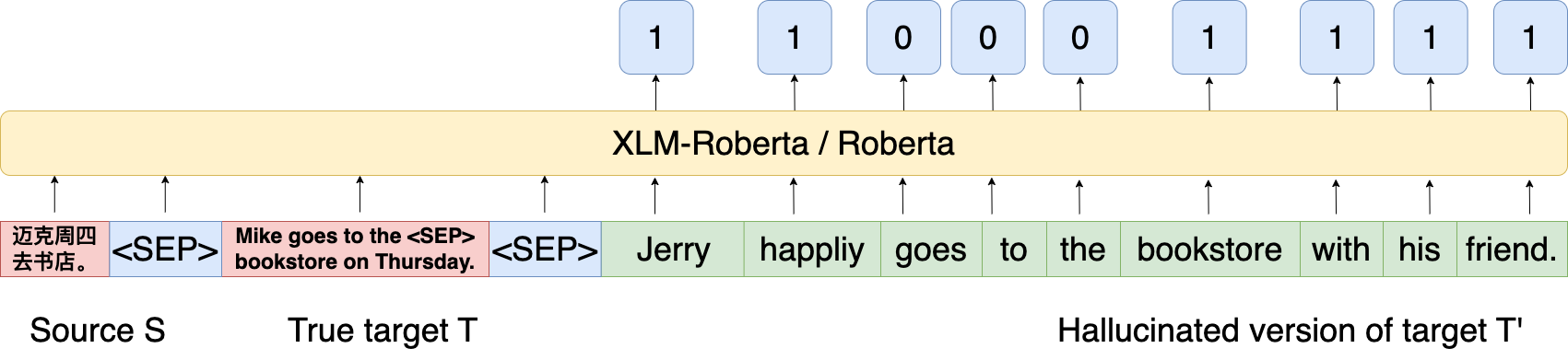}
    \caption{Finetuning XLM-Roberta (for cross-lingual generation task, e.g. MT) or Roberta (for monolingual generation task, e.g. text summarization) on the synthetic training data.~(Figure source:~\cite{zhou2021detecting})}
    \label{fig:zhou_finetune}
\end{figure}

Methodology-wise, the authors propose a general-purpose method for token-level hallucination detection for conditional sequence generation tasks. Given the source input, the authors formulate the task of token-level hallucination detection as a sequence labeling problem where a binary label is predicted at each position of the machine generation. The authors finetune a cross-lingual language model (LM)~\cite{conneau2020unsupervised} for MT and a monolingual LM~\cite{zhuang-etal-2021-robustly} for summarization. In both cases, the input consists of concatenating the original text, the true target (translation or summary), and the hallucinated version of the target as a single input sequence to the model. Then the standard classification loss is used to to carry out the finetuning process. A graphical illustration of this procedure can be found in Figure~\ref{fig:zhou_finetune}. 

\begin{table}[t]
\centering
\caption{F1 (x100) of hallucination labels on MT and abstractive summarization tasks. The first block are baseline methods and the second block are reported results of the proposed method. Bold indicates best results not using references.~(Table source:~\cite{zhou2021detecting})}
\label{tab:zhou_result}
\resizebox{\textwidth}{!}{%
\begin{tabular}{l|ll|llll}
\hline
\multirow{2}{*}{Methods} & \multicolumn{2}{c|}{MT} & \multicolumn{4}{c}{Summarization} \\
 & \multicolumn{1}{c}{TranS2S} & \multicolumn{1}{c|}{MBART} & \multicolumn{1}{c}{PtGen} & \multicolumn{1}{c}{TConvS2S} & \multicolumn{1}{c}{TranS2S} & \multicolumn{1}{c}{BERTS2S} \\ \hline
Alignment & 29.47 & 9.93 & 38.92 & 37.94 & 34.37 & 25.81 \\
Overlap-based & 9.14 & 3.24 & 57.22 & 54.25 & 53.79 & 55.13 \\
Synonym-based & - & - & 59.54 & 63.73 & 58.66 & 53.07 \\ \hline
Ours (w/o reference) & 65.75 & 41.92 & 63.66 & 65.94 & 61.70 & 55.45 \\
Ours (w/o reference + synonym) & - & - & \textbf{64.72} & \textbf{69.37} & \textbf{63.88} & \textbf{56.49} \\
Ours (w/ reference) & \textbf{66.08} & \textbf{46.81} & 63.89 & 66.28 & 62.24 & 55.88 \\ \hline
\end{tabular}%
}
\end{table}

Table~\ref{tab:zhou_result} shows the F1 scores of token-level hallucination labels across six benchmark datasets for MT and abstractive summarization. The proposed method achieves decent performance on this task and ranks the best among all baseline methods. One thing worth mentioning is that, during training, the true target is concatenated into the input of the model. However, during the testing phase, due to the lack of the true target, the authors propose to replace this part of input by some dummy variables. Surprisingly, the proposed model can generalize well without references. As a contrast, the authors further report that the model can achieve a significantly higher recall but worse precision when including the true target as a part of the input at test time.

\subsection{Sentence Level}
\subsubsection[Selfcheckgpt: Zero-resource black-box hallucination detection for generative large language models]{Selfcheckgpt: Zero-resource black-box hallucination detection for generative large language models~\cite{manakul2023selfcheckgpt}}
\label{sec:selfcheck}
\textbf{Github Link: } https://github.com/potsawee/selfcheckgpt
\newline
\textbf{Tasks: } Question and answering.
\newline
\textbf{Core Idea: } This paper makes a claim that most of the hallucinations happen when the model is unsure about itself. Therefore, for this type of hallucinations, one can detect it via evaluating the LLM's self-consistency by checking the similarity between the main response and a set of sample responded generated with the same LLM with higher temperature. 
\newline
\newline
\begin{figure}[h]
    \centering
    \includegraphics[width=0.4\linewidth]{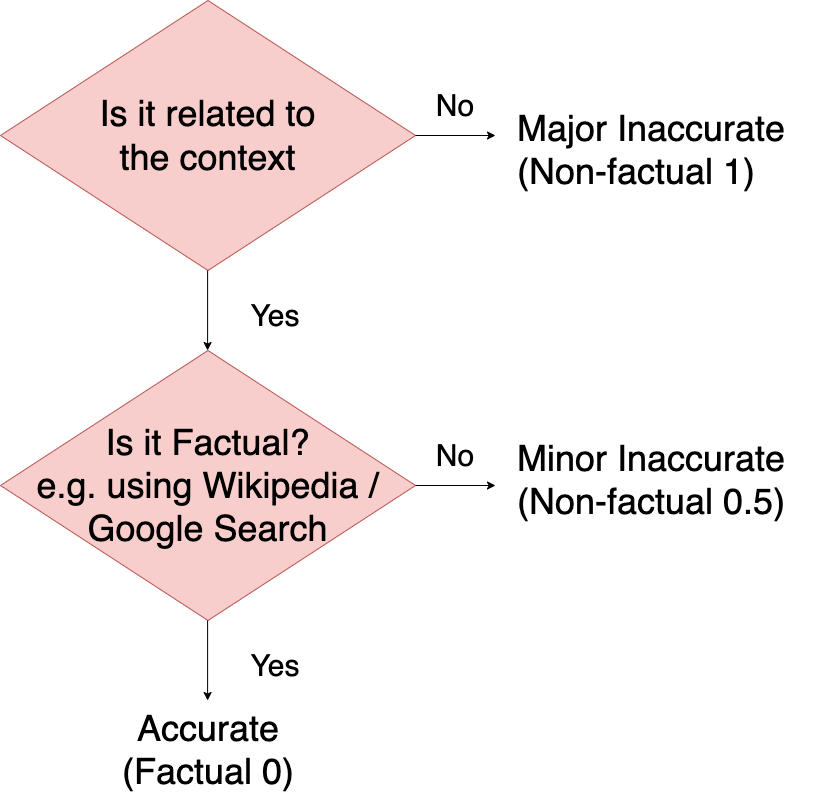}
    \caption{Flowchart of the data annotation process.~(Figure source:~\cite{manakul2023selfcheckgpt})}
    \label{fig:selfcheck_annotation}
\end{figure}

In this study, the authors introduce "SelfCheckGPT," a straightforward sampling-based approach designed for fact-checking large language models (LLMs) in a zero-resource manner, i.e., without relying on an external database. SelfCheckGPT capitalizes on the intuitive notion that if an LLM possesses knowledge of a particular concept, sampled responses are likely to be consistent and contain similar facts. Conversely, for hallucinated facts, stochastically sampled responses are expected to diverge and contradict each other. By extracting multiple responses from an LLM, one can assess information consistency among the different responses, thereby discerning factual statements from hallucinated ones.

The paper explores three variants of SelfCheckGPT for measuring informational consistency: BERTScore ~\cite{bert-score}, question-answering ~\cite{manakul2023mqag}, and n-grams. Additionally, they also delve into natural language inference scores using DeBERTa-v3-large~\cite{he2022debertav3} and prompt engineering using LLMs. The authors validate this approach using GPT-3~\cite{brown2020language} to generate passages from the WikiBio dataset~\cite{lebret-etal-2016-neural} about each individual topics, then manually annotate the factuality of the generated passages using human judges. 
The study demonstrates that SelfCheckGPT can i) identify non-factual and factual sentences and ii) rank passages in terms of factuality. Empirical comparisons underscore that, in sentence-level hallucination detection tasks, the proposed approach outperforms various baseline methods.

\begin{figure}[t]
    \centering
    \includegraphics[width=0.8\linewidth]{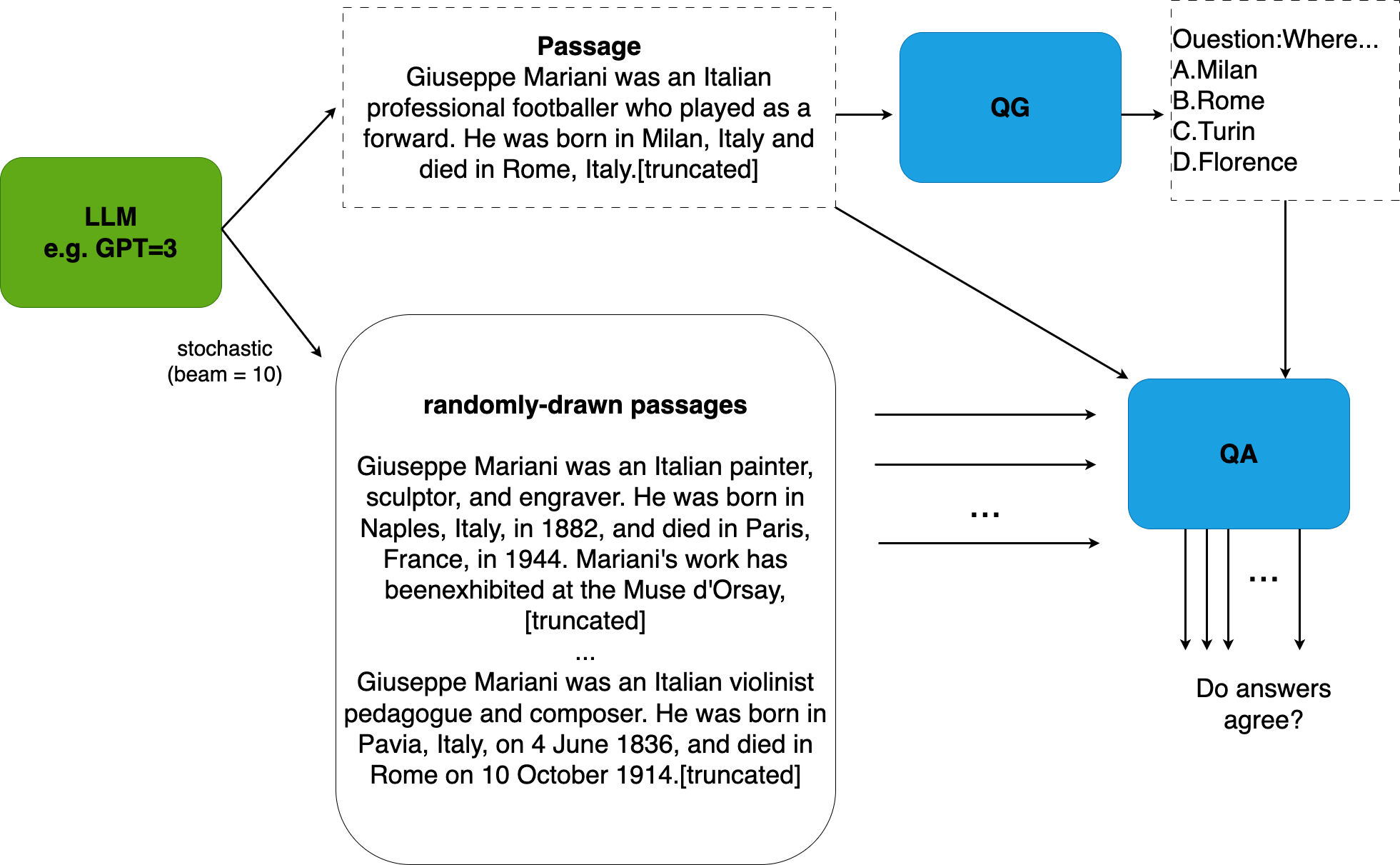}
    \caption{SelfCheckGPT with Question Answering.~(Figure source:~\cite{manakul2023selfcheckgpt})}
    \label{fig:selfcheck_procedure}
\end{figure}

For the dataset, the authors opt to use WikiBio~\cite{lebret-etal-2016-neural} as a starting point. WikiBio is a dataset of the first paragraph (with tabular information) of Wikipedia articles about specific concepts. The goal of this dataset is to generate passages using LLMs and evaluate the factuality with respect to the original article. To do so,  they rank the WikiBio test set in terms of paragraph length and randomly sample 238 articles from the top $20\%$ of the longest articles (to ensure no obscure concept is selected). Then, they use GPT-3 (text-davinci-003)~\cite{brown2020language} to generate articles on a specific concept in the dataset, using the prompt "This is a Wikipedia passage about \{concept\}". There are in total 238 passages which contains 1,908 sentences. Then the authors ask human judges to annotate each of the generated sentences into three classes: major inaccurate, minor inaccurate and accurate, based on the process described in Figure~\ref{fig:selfcheck_annotation}. Of the 1,908 annotated sentences, 761 ($39.9\%$) of the sentences are labelled major-inaccurate, 631 ($33.1\%$) are minor-inaccurate, and 516 ($27.0\%$) are accurate.

The idea of SelfCheckGPT is to randomly sample various responses and compute a consistency score between these generated samples and the main response. For the target LLM, one should generate the main response passage using the regular setting and then generate $n$ number of randomly-drawn passages using the same query as the list of references. To ensure the randomness, one common practice is to increase the temperature parameter of the LLM. Then a consistency score will be computed between the main response passage and the randomly-drawn passages. The scoring can be done in various ways as mentioned in the previous paragraph, such as question answering or n-grams, etc. The process of SelfCheckGPT using question answering~\cite{manakul2023mqag} as the consistency score can be found in Figure~\ref{fig:selfcheck_procedure}.

Experiments are conducted to determine whether SelfCheckGPT is capable of identifying the factuality of sentences. In detecting non-factual sentences, both major-inaccurate labels and minor-inaccurate labels are grouped together into the non-factual class, while the factual class refers to accurate sentences. Table~\ref{tab:selfcheck_result} shows correlation results (higher the better) with respect to human judges on the sentence-level hallucination detection task. The results in Table~\ref{tab:selfcheck_result} show that all of the SelfCheckGPT methods correlate the best with human judgements among all the compared baseline methods. Further, the three variants of SelfCheckGPT appear complementary, with the combined approach being the best-performing system, achieving the highest Pearson correlation of 69.05. 

\begin{table}[t]
\centering
\caption{AUC-PR for sentence-level detection tasks. Passage-level ranking performances are measured by Pearson correlation coefficient and Spearman’s rank correlation coefficient w.r.t. human judgements.~(Table source:~\cite{manakul2023selfcheckgpt})}
\label{tab:selfcheck_result}
\resizebox{\textwidth}{!}{%
\begin{tabular}{llllll}
\hline
\multicolumn{1}{l|}{} & \multicolumn{3}{l|}{Sentence-level (AUC-PR)} & \multicolumn{2}{l}{Passage-level (Corr.)} \\
\multicolumn{1}{l|}{\multirow{-2}{*}{Method}} & NonFact & NonFact* & \multicolumn{1}{l|}{Factual} & Pearson &Spearman \\ \hline
\multicolumn{1}{l|}{Random} & 72.96 & 29.72 & \multicolumn{1}{l|}{27.04} & - & - \\ \hline
\multicolumn{6}{l}{GPT-3's probabilities (LLM, grey-box)} \\ \hline
\multicolumn{1}{l|}{Avg(-logp)} & 83.21 & 38.89 & \multicolumn{1}{l|}{53.97} & 57.04 & 53.93\\
\multicolumn{1}{l|}{Avg(H)+} & 80.73 & 37.09 & \multicolumn{1}{l|}{52.07} & 55.52 &50.87 \\
\multicolumn{1}{l|}{Max(-logp)} & 87.51 & 35.88 & \multicolumn{1}{l|}{50.46} & 57.83 & 55.69 \\
\multicolumn{1}{l|}{Max(H)} & 85.75 & 32.43 & \multicolumn{1}{l|}{50.27} & 52.48 & 49.55 \\ \hline
\multicolumn{6}{l}{LLaMA-30B's probabilities (Proxy LLM, black-box)} \\ \hline
\multicolumn{1}{l|}{Avg(-logp)} & 75.43 & 30.32 & \multicolumn{1}{l|}{41.29} & 21.72 & 20.20 \\
\multicolumn{1}{l|}{Avg(H)} & 80.80 & 39.01 & \multicolumn{1}{l|}{42.97} & 33.80 & 39.49 \\
\multicolumn{1}{l|}{Max(-logp)} & 74.01 & 27.14 & \multicolumn{1}{l|}{31.08} & -22.83 & -22.71 \\
\multicolumn{1}{l|}{Max(H)} & 80.92 & 37.32 & \multicolumn{1}{l|}{37.90} & 35.57 & 38.94 \\ \hline
\multicolumn{6}{l}{SelfCheckGPT (black-box)} \\ \hline
\multicolumn{1}{l|}{w/ BERTScore} & 81.96 & 45.96 & \multicolumn{1}{l|}{44.23} & 58.18 & 55.90 \\
\multicolumn{1}{l|}{w/ QA} & 84.26 & 40.06 & \multicolumn{1}{l|}{48.14} & 61.07 & 59.29 \\
\multicolumn{1}{l|}{w/ Unigram (max)} & 85.63 & 41.04 & \multicolumn{1}{l|}{58.47} & 64.71 & 64.91 \\
\multicolumn{1}{l|}{Combination} & 87.33 & 44.37 & \multicolumn{1}{l|}{61.83} & 69.05 & 67.77 \\ \hline
\end{tabular}%
}
\end{table}

\subsubsection[ALIGNSCORE: Evaluating Factual Consistency with A Unified Alignment Function]{ALIGNSCORE: Evaluating Factual Consistency with A Unified Alignment Function~\cite{zha-etal-2023-alignscore}}
\textbf{Github Link: } https://github.com/yuh-zha/AlignScore
\newline
\textbf{Tasks: } NLI, QA, paraphrasing, fact verification, information retrieval, semantic similarity, and summarization.
\newline
\textbf{Core Idea: } This paper proposes a new general factual consistency metric based on an unified text-to-text information alignment function through unifying a wide range of data sources and NLG tasks. 
\newline
\newline
Previous research, exemplified by the aforementioned SelfCheckGPT, has developed several metrics primarily reliant on specific functions such as natural language inference (NLI) or question answering (QA). However, these metrics are typically trained on limited data and are consequently confined to assessing a singular type of task, often overlooking diverse factual inconsistencies (e.g., contradictions, hallucinations) that manifest in various inputs/outputs (e.g., sentences, documents) across different tasks.

This paper introduces ALIGNSCORE as a novel, comprehensive factual consistency metric founded on a unified text-to-text information alignment function. The authors propose a model that unifies a broad spectrum of data sources, leveraging massive and diverse datasets to train a general information alignment model. This model estimates an alignment score when presented with two arbitrary pieces of text. To achieve this, the authors reformat and aggregate data from 15 datasets spanning 7 prominent language tasks, including NLI, QA, paraphrasing, fact verification, information retrieval, semantic similarity, and summarization. This approach yields a total of 4.7 million training examples with diverse characteristics, resulting in an alignment function characterized by remarkable generalizability.

ALIGNSCORE is constructed using the alignment function as a foundational element. The authors conduct extensive experiments on large-scale benchmarks encompassing 22 evaluation datasets, with 19 of these datasets not encountered during the alignment training. Impressively, ALIGNSCORE demonstrates significant improvement over a broad array of previous metrics. Furthermore, with 355 million parameters, ALIGNSCORE either matches or surpasses metrics based on ChatGPT and GPT-4, despite the latter being orders of magnitude larger.

Methodology-wise, they propose the unified alignment functions.  They first train the alignment function by unifying a large diversity of data sources. Then they define ALIGNSCORE by combining the alignment function with a new context/claim splitting and aggregation strategy. 

\begin{figure}[t]
    \centering
    \includegraphics[width=0.9\linewidth]{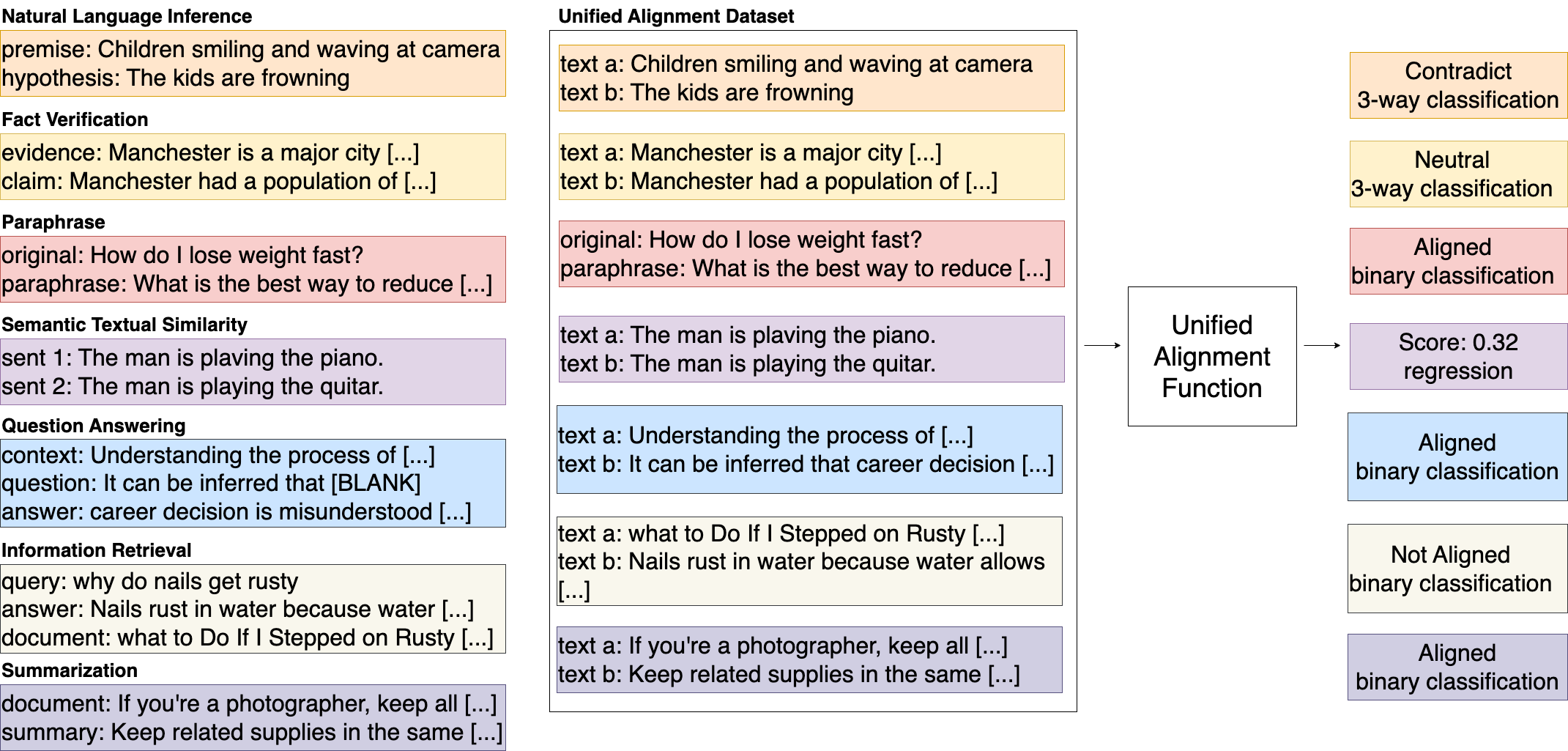}
    \caption{The information alignment problem and how we unify various tasks into the alignment task. We convert each sample in the tasks we consider into a text pair ($a$, $b$), and the alignment function predicts a label $y$ characterizing the level of alignment. The underlined text indicates items in the original dataset (e.g., question and answer in a QA dataset) are combined to form part of the text pair in the alignment dataset.~(Figure source: ~\cite{zha-etal-2023-alignscore})}
    \label{fig:alignscore_alignment}
\end{figure}

The idea of the ALIGNSCORE comes from the so-called alignment function. Given two pieces of text $a$ and $b$, $b$ is considered to be aligned with $a$ if all information in $b$ is present in $a$ and does not contradict $a$. A typical set-up is that $b$ is a claim that the LLM makes (answer) while $a$ is the context provided to the LLM (query). Ideally, we want the claim to be supported by the context. Conceptually, one can model the information alignment as a function that maps the text pair $(a, b)$ to a label $y$ that characterizes the level of alignment, i.e., $f(a, b) = y$, which is defined to be the alignment function. 

In order to train the alignment function, the authors opt to adapt and aggregate diverse language tasks to form an unified alignment training corpus (Figure~\ref{fig:alignscore_alignment}). One challenge of learning across multiple data sources and NLG tasks is to unify the input and output. To unify input formats, the authors convert each sample into a text pair $(a, b)$, with specific treatments for some irregular tasks, such as QA. To unify output formats, the authors convert all tasks into a set of related alignment problems to preserve as much information as possible from the original datasets. To do so, they devise three options for the alignment label $y$:
\begin{align*}
    y_{bin} &\in \{ALIGNED, NOT-ALIGNED\},\\
    y_{3way} &\in \{ALIGNED, CONTRADICT, NEUTRAL\},\\
    y_{reg} &\in [0,1].
\end{align*}
More concretely, for tasks that come with discrete labels, depending on their setup, the alignment function predicts either the binary classification label $Y_{bin}$ (paraphrase, QA, information retrieval, and summarization) or the 3-way classification label $y_{3way}$ (NLI, and fact verification); for tasks with continuous labels (semantic textual similarity), the alignment function predicts the regression label $y_{reg}$. Here a higher $y_{reg}$ indicates that more information in $b$ is supported by $a$. The authors build the alignment model with a language model (e.g., RoBERTa~\cite{zhuang-etal-2021-robustly}) and 3 individual linear layers as the 3-way classification ($y_{3way}$), binary classification ($y_{bin}$), and regression ($y_{reg}$) heads. 

\begin{table}[t]
\centering
\caption{The AUC-ROC of different metrics on the SummaC benchmark. The last column (AVG) is the average performance of each metric. The dark green indicates the best metric on each dataset or on average. And the light green indicates the second best. CGS and XSF are abbreviations for CoGenSumm and XSumFaith, respectively.~(Table source: ~\cite{zha-etal-2023-alignscore})}
\label{tab:alignscore_result}
\resizebox{\textwidth}{!}{%
\begin{tabular}{l|llllllll}
\hline
Type & Metric & CGS & XSF & PolyTope & FactCC & SummEval & FRANK & AVG \\ \hline
\multirow{3}{*}{QA} & FEQA & 53.7 & 47.6 & 54.3 & 47.9 & 48.8 & 37.2 & 48.3 \\
 & QuestEval & 60.4 & 63.6 & 77.0 & 74.2 & 74.3 & 85.8 & 72.5 \\
 & QAFactEval & 83.4 & 66.1 & 86.4 & 89.2 & 88.1 & 89.4 & 83.8 \\ \hline
\multirow{7}{*}{\begin{tabular}[c]{@{}l@{}}Similarity\\ Matching\end{tabular}} & ROUGE-1 & 69.7 & 64.5 & 82.5 & 75.8 & 87.2 & 85.0 & 77.4 \\
 & ROUGE-2 & 70.5 & 65.9 & 83.7 & 76.0 & 87.2 & 85.3 & 78.1 \\
 & ROUGE-L & 70.2 & 62.9 & 81.9 & 76.3 & 87.3 & 85.3 & 77.3 \\
 & BLEU & 71.8 & 55.8 & 86.9 & 75.0 & 83.8 & 84.5 & 76.3 \\
 & BERTScore & 63.1 & 49.0 & 85.3 & 70.9 & 79.6 & 84.9 & 72.1 \\
 & NER-Overlap & 51.1 & 64.9 & 72.1 & 49.8 & 56.6 & 68.1 & 60.4 \\
 & SimCSE & 56.2 & 62.2 & 75.2 & 59.0 & 77.2 & 74.8 & 67.4 \\ \hline
Regression & BLEURT & 60.8 & 64.7 & 76.7 & 59.7 & 71.1 & 82.5 & 69.2 \\ \hline
\multirow{4}{*}{NLI} & MNLI & 44.9 & 46.6 & 45.0 & 48.3 & 43.5 & 59.3 & 47.9 \\
 & DAE & 52.4 & 76.7 & 72.8 & 54.2 & 66.1 & 78.9 & 66.8 \\
 & SummaC-ZS & 73.6 & 58.0 & 87.5 & 83.7 & 85.8 & 85.3 & 79.0 \\
 & SummaC-CONV & 67.2 & 70.3 & 81.8 & 92.3 & 86.1 & 88.5 & 81.0 \\ \hline
\multirow{5}{*}{Misc} & UniEval & 84.7 & 65.5 & \textbf{93.4} & 89.9 & 86.3 & 88.0 & 84.6 \\
 & CTC & 76.5 & 65.9 & 89.5 & 82.6 & 85.6 & 87.3 & 81.2 \\
 & BARTScore & 74.3 & 62.6 & 91.7 & 82.3 & 85.9 & 88.5 & 80.9 \\
 & FactCC & 64.9 & 55.1 & 78.5 & 72.7 & 71.8 & 69.8 & 68.8 \\
 & BLANC & 54.1 & 53.5 & 74.7 & 56.4 & 68.6 & 83.4 & 65.1 \\ \hline
\multirow{2}{*}{Ours} & ALIGNSCORE-base & 83.7 & \textbf{79.4} & 87.8 & 93.3 & 89.9 & 90.5 & 87.4 \\
 & ALIGNSCORE-large & \textbf{86.4} & 75.8 & 92.4 & \textbf{93.7} & \textbf{91.7} & \textbf{91.4} & \textbf{88.6} \\ \hline
\end{tabular}%
}
\end{table}

Then author then propose to compute the sentence level ALIGNSCORE using the result from the alignment function. Specifically, for each sentence in the claim passage, the evaluation is done against all context sentences using the alignment function. Then,the highest alignment score is selected for each claim sentence as the sentence level ALIGNSCORE. In addition, one can obtain the document-level ALIGNSCORE by averaging all the sentence level ones. Note this method is quite similar to SelfCheckGPT in Section~\ref{sec:selfcheck}. The difference here is that ALIGNSCORE assumes the context passage is given, while in SelfCheckGPT obtain the context by random generation with high-temperature LLM. Here one remark is that we can potentially combine the two methods, using ALIGNSCORE as a consistency score to plug in the SelfCheckGPT framework.

The experiments of this work focus more on the summarization. The result shown in Table~\ref{tab:alignscore_result} is based on the SummaC dataset~\cite{laban-etal-2022-summac}, which standardizes the task of summary inconsistency detection by casting it as a binary classification problem. Here the context would be the source document, while the claim is the summary generated by the LLM. One can observe that ALIGNSCORE-large achieves the best average performance on the SummaC benchmark, scoring the highest in 4 out of 6 datasets. Recall the authors use RoBERTa as a base model to finetune the alignment function. For ALIGNSCORE-large in the table, it leverages RoBERTa-large as the base model and ALIGNSCORE-base uses the basic RoBERTa model as the base model.


\subsubsection["Why is this misleading?": Detecting News Headline Hallucinations with Explanations]{"Why is this misleading?": Detecting News Headline Hallucinations with Explanations~\cite{shen2023misleading}}
\textbf{Github Link: } No code provided.
\newline
\textbf{Dataset Link: } https://bit.ly/exhalder-dataset
\newline
\textbf{Tasks: } Summarization.
\newline
\textbf{Core Idea: } This paper proposes a new framework to address hallucination detection under the task of headline generation. Exhalder leverages an explainer that adapts the knowledge from public natural language inference datasets into the news domain, to further improve the hallucination detection result. 
\newline
\newline

\begin{figure}[t]
    \centering
    \includegraphics[width=0.9\linewidth]{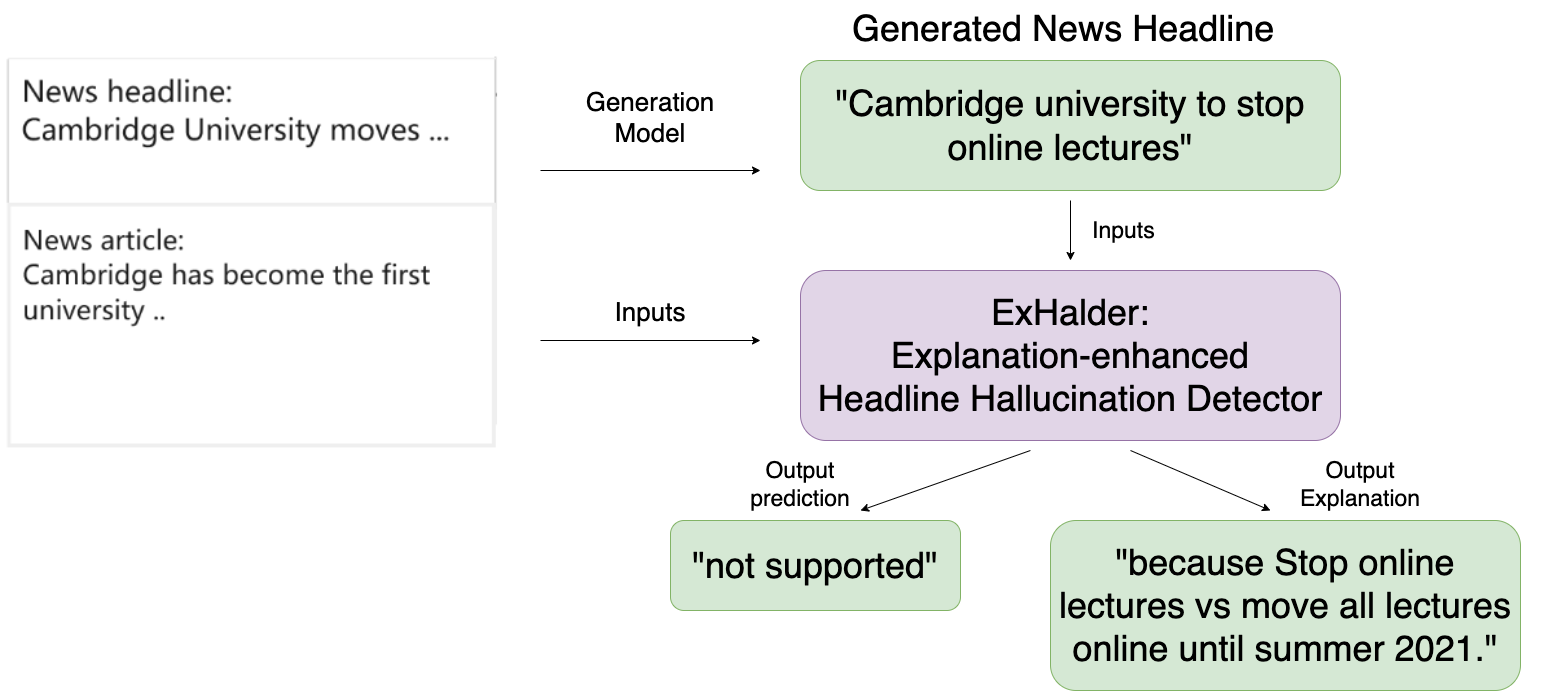}
    \caption{An illustrative example of automated news head-
line hallucination detection with a model generated natural
language explanation.~(Figure source:~\cite{shen2023misleading})}
    \label{fig:exhalder_overview}
\end{figure}

In this work, the authors propose a new framework named ExHalder to address this challenge for headline hallucination detection. 
Automatic news headlines generation, often viewed as a specialized document summarization task, focuses on creating headline-style summaries for news articles.
Its significance has grown in rapidly conveying ongoing news to users, becoming a pivotal feature in the news domain. LLM-based approaches, especially in automatic headline generation, has become widely favored. However, addressing hallucination issues poses a critical challenge for deploying this feature in web-scale systems, demanding factual accuracy in the news domain. The proposed method integrates knowledge from a public natural language inference dataset into the news domain, generating natural language sentences to elucidate hallucination detection results. 
Figure~\ref{fig:exhalder_overview} shows an example of the hallucination detection pipeline leveraging ExHalder. To evaluate the model performance, the authors collect a dataset with more than six thousand labeled (article, headline) pairs. Extensive experiments on this dataset and another six public ones demonstrate that ExHalder can identify hallucinated headlines accurately and justifies its predictions with human-readable natural language explanations.

\begin{figure}[t]
    \centering
    \includegraphics[width=\linewidth]{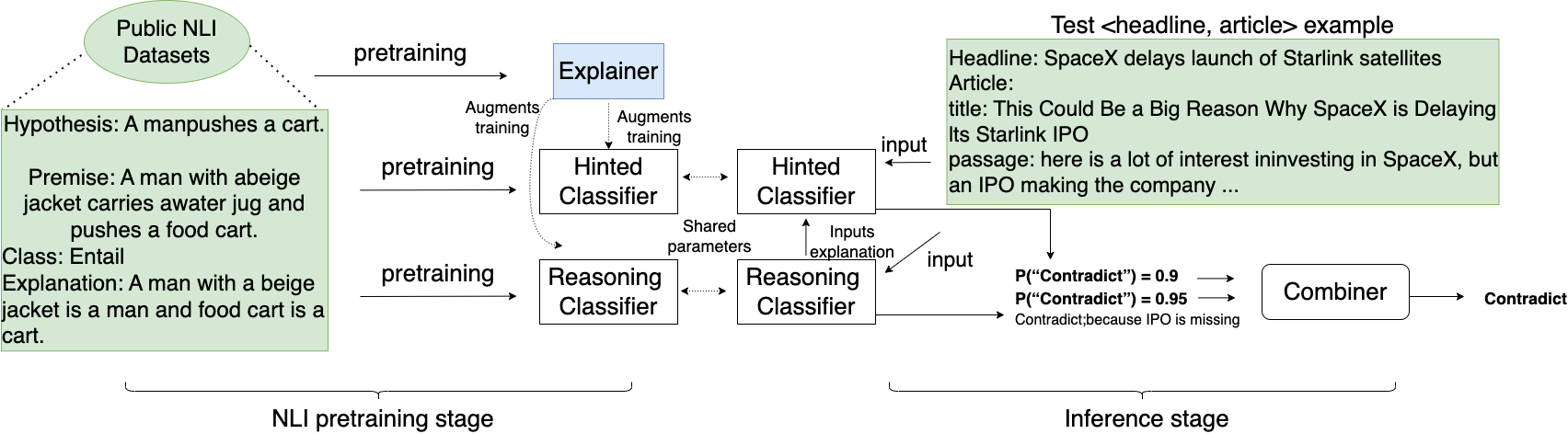}
    \caption{The ExHalder framework overview.~(Figure source:~\cite{shen2023misleading})}
    \label{fig:exhalder_framework}
\end{figure}

The ExHalder framework is built upon three key components for news headline hallucination detection: (1) a reasoning classifier which 
takes (article, headline) pairs as inputs and examine if the headline is contradicting the article and outputs the class label with the explanation of the classification; (2) a hinted classifier which receives as input the (article, headline, explanation) triplet and predicts the if the headline is hallucinated, where the explanation acts as a "hint" in this classifier, providing some extra information; (3) an explainer that generates the natural language explanation based on the input (article, headline) with its known class label. Then, in order to avoid the data scarcity issue on labeling, the authors propose to (1) pretrain all the three components with large-scale natural language inference (NLI) datasets, and (2) use augmented training with the explainer to further finetune the hinted-classifier and the reasoning classifier. Figure~\ref{fig:exhalder_framework} shows an overview of the ExHalder framework.

In this paper, the authors collect a new dataset that contains 6,270 human curated examples, where each example consists of a triplet of (news article, news headline, hallucination label). Among these data, they split 5,190 examples for training, 349 examples for validation, and 731 examples for testing. The headline is generated from NHNet~\cite{gu2020generating} and the label is obtained from multiple human experts according to a common guideline. Specifically, three full-time journalism degree holders were involved to determine the final hallucination label of each example via majority voting. Among all the examples, 1,934 of them are labeled as "hallucinated" and the remaining 4,336 examples are labeled as "entailed" (not hallucinated). Furthermore, there are 2,074 examples with additional rater-written comments (besides binary hallucination labels) and they treat them as user-provided explanations. 

Table~\ref{tab:exhalder_result} shows the qualitative results of ExHalder and other baseline methods. Some explanations of the abbreviations:
\begin{itemize}
    \item ExHalder-NoPT: ExHalder without the NLI-based pretraining step. 
    \item ExHalder-NoEX: ExHalder with the NLI-based pretraining step but without leveraging any explanation information. 
    \item ExHalder-NoHC: ExHalder without the hinted classifier module.
\end{itemize}

In the findings, firstly, it is evident that traditional methods employing manual feature engineering (e.g., SVM, XGBoost) yield subpar results, underscoring the challenge of headline hallucination detection. This difficulty highlights the necessity for models capable of capturing nuanced semantic distinctions between the article and its headline.
Secondly, the comparison between ExHalder and ExHalder-NoPT reveals that the NLI-based pretraining significantly enhances the identification of hallucinated headlines. This improvement is attributed to the model being primed with entailment task semantics.
Thirdly, the performance gap between ExHalder and ExHalder-NoEX indicates additional advancements can be achieved by injecting the explanation information into the model
training process. 
Finally, ExHalder demonstrates the most superior performance across all metrics, outperforming the second-best method by a substantial margin.

\begin{table}[t]
\centering
\caption{Quantitative results on the news headline hallucination detection dataset. The superscript $^{*}$ means the improvement is statistically significant compared to $T5_{xxl}$.~(Table source:~\cite{shen2023misleading})}
\label{tab:exhalder_result}
\resizebox{0.6\textwidth}{!}{%
\begin{tabular}{l|llll}
\hline
Methods & Accuracy & Precision & Recall & F1 \\ \hline
SVM & 57.31 & 28.65 & 20.50 & 23.90 \\
XGBoost & 60.19 & 42.39 & 60.67 & 49.91 \\ \hline
BERT-base & 73.46 & 71.43 & 31.38 & 43.60 \\
T5-xxl & 82.39 & 76.29 & 66.93 & 71.29 \\
T5-xxl+EXP & 82.62 & 78.98 & 64.15 & 70.63 \\ \hline
ExHalder-NoPT & 82.08 & 75.96 & 66.11 & 70.69 \\
ExHalder-NoEX & 83.17 & 80.01 & 64.71 & 71.54 \\
ExHalder-NoHC & 84.08 & 82.06 & 65.69 & 72.96 \\
ExHalder & \textbf{84.46} & \textbf{82.63} & \textbf{67.16} & \textbf{74.08} \\ \hline
\end{tabular}%
}
\end{table}

\subsubsection[HaRiM+: Evaluating Summary Quality with Hallucination Risk]{HaRiM+: Evaluating Summary Quality with Hallucination Risk~\cite{son-etal-2022-harim}}
\textbf{Huggingface API: } https://huggingface.co/spaces/NCSOFT/harim$\_$plus
\newline
\textbf{Tasks: } Summarization.
\newline
\textbf{Core Idea: } This paper repurposes the decoder overconfidence-regularizing objective as a hallucination risks measurement and propose a hallucination detection metric, ready to be used as a Huggingface API.
\newline
\newline

In this study, the authors reinterpret the decoder overconfidence-regularizing objective suggested in~\cite{miao2021prevent} as a hallucination risk measurement to better estimate the quality of generated summaries. The authors propose a reference-free metric, HaRiM+ for summarization tasks. The metric only requires an off-the-shelf summarization model to compute the hallucination risk based on token likelihoods. The metric is very stright-forward to deploy, with no additional training or human-alignment. To evaluate its performance for summarization, the authors compare the proposed metric with various baselines on three annotated datasets: FRANK~\cite{pagnoni-etal-2021-understanding}, QAGs~\cite{wang-etal-2020-asking} and SummEval~\cite{fabbri2020summeval}, where Harim+ shows state-of-the-art correlation to human judgment. 

It is a commonly-known problem where the decoder relies too much on the decoder's own context while being less dependent on the encoder's counterpart in a encoder-decoder architecture~\cite{bowman2016generating}.~\cite{miao2021prevent} introduced a margin-based token-level objective as a regularization term that prevents the decoder from focusing too much on the decoder-side context. 
Under summarization task, hallucination is often considered to be not faithful to the source document, which can be seen as the encoder's context. Therefore, one can reinterpret the aforementioned regularization term as the hallucination risk, which is the core idea of Harim+. 
Harim+ is defined over the similar concepts of prior and posterior probabilities we mentioned in the subsection~\ref{sec:enfa}, where the prior probability is defined as the probability of its generation by a language model that does not have access to the source text, while the posterior probability is computed by conditioning on the source document. 

\begin{table}
\centering
\caption{Metric-to-human judgement correlation (segment level) reported in Kendall's $\tau$ . Bold-face values are the largest correlating metrics, underlined are second-large values amongst the metrics. HaRiM+ outperforms others in most criteria. SummEval's quality criteria; consistency, coherence, fluency, and relevance are abbreviated as Con, Coh, Flu, and Rel respectively.~(Table source:~\cite{son-etal-2022-harim})}
\label{tab:harim_result}
\resizebox{\textwidth}{!}{%
\begin{tabular}{lllllllll}
\hline
\multicolumn{1}{l|}{\textbf{}} & \multicolumn{6}{c|}{CNNDM} & \multicolumn{2}{l}{XSUM} \\ \hline
\multicolumn{1}{l|}{Kendall's $\tau$} & FRANK & QAGS & \multicolumn{4}{l|}{SummEval} & FRANK & QAGS \\ \hline
\multicolumn{1}{l|}{Metrics} & Factuality & Factuality & Con & Coh & Flu & \multicolumn{1}{l|}{Rel} & Factuality & Factuality \\ \hline
\multicolumn{9}{l}{N - gram - matching} \\ \hline
\multicolumn{1}{l|}{ROUGE 1} & 0.182 & -0.052 & 0.105 & 0.123 & 0.062 & \multicolumn{1}{l|}{0.209} & 0.125 & 0.110 \\
\multicolumn{1}{l|}{ROUGE 2} & 0.135 & -0.107 & 0.101 & 0.097 & 0.048 & \multicolumn{1}{l|}{0.153} & 0.128 & 0.097 \\
\multicolumn{1}{l|}{ROUGE L} & 0.141 & -0.072 & 0.091 & 0.113 & 0.061 & \multicolumn{1}{l|}{0.164} & 0.117 & 0.090 \\
\multicolumn{1}{l|}{METEOR} & 0.198 & 0.053 & 0.125 & 0.116 & 0.070 & \multicolumn{1}{l|}{0.223} & 0.121 & 0.115 \\
\multicolumn{1}{l|}{sacreBLEU} & 0.136 & -0.085 & 0.080 & 0.167 & 0.088 & \multicolumn{1}{l|}{0.131} & 0.113 & 0.012 \\ \hline
\multicolumn{1}{l|}{ROUGE 1\_art} & 0.185 & 0.243 & 0.111 & 0.036 & 0.058 & \multicolumn{1}{l|}{0.127} & -0.003 & -0.074 \\
\multicolumn{1}{l|}{ROUGE 2\_art} & 0.249 & 0.315 & 0.195 & 0.072 & 0.119 & \multicolumn{1}{l|}{0.165} & 0.027 & 0.069 \\
\multicolumn{1}{l|}{ROUGE L\_art} & 0.225 & 0.305 & 0.203 & 0.097 & 0.123 & \multicolumn{1}{l|}{0.050} & 0.010 & -0.019 \\
\multicolumn{1}{l|}{METEOR\_art} & 0.174 & 0.234 & 0.112 & 0.009 & 0.071 & \multicolumn{1}{l|}{0.091} & 0.004 & -0.052 \\
\multicolumn{1}{l|}{sacreBLEU\_art} & 0.153 & 0.245 & 0.091 & 0.042 & 0.035 & \multicolumn{1}{l|}{} & -0.038 & -0.139 \\ \hline
\multicolumn{9}{l}{N - gram stats} \\ \hline
\multicolumn{1}{l|}{NovelNgram\_4} & 0.275 & 0.392 & 0.221 & 0.203 & 0.173 & \multicolumn{1}{l|}{0.205} & 0.017 & 0.056 \\
\multicolumn{1}{l|}{NovelNgram\_3} & 0.273 & 0.370 & 0.218 & 0.208 & 0.171 & \multicolumn{1}{l|}{0.208} & 0.064 & 0.080 \\
\multicolumn{1}{l|}{NovelNgram\_2} & 0.259 & 0.327 & 0.199 & 0.209 & 0.150 & \multicolumn{1}{l|}{0.207} & 0.053 & 0.129 \\
\multicolumn{1}{l|}{NovelNgram\_1} & 0.219 & 0.201 & 0.090 & 0.190 & 0.068 & \multicolumn{1}{l|}{0.173} & 0.091 & 0.120 \\
\multicolumn{1}{l|}{Length ( no . tokens )} & 0.187 & 0.185 & 0.078 & 0.033 & 0.000 & \multicolumn{1}{l|}{0.000} & -0.111 & -0.132 \\ \hline
\multicolumn{9}{l}{Contextual Embedding} \\ \hline
\multicolumn{1}{l|}{BERTScore P} & 0.168 & -0.067 & 0.041 & 0.229 & 0.097 & \multicolumn{1}{l|}{0.192} & \textbf{0.151} & 0.016 \\
\multicolumn{1}{l|}{BERTScore R} & 0.250 & 0.017 & 0.125 & 0.241 & 0.097 & \multicolumn{1}{l|}{0.299} & 0.107 & 0.058 \\
\multicolumn{1}{l|}{BERTScore F1} & 0.232 & -0.029 & 0.079 & 0.267 & 0.111 & \multicolumn{1}{l|}{0.267} & 0.142 & 0.036 \\ \hline
\multicolumn{1}{l|}{BERTScore P\_art} & 0.301 & 0.331 & 0.266 & 0.308 & 0.236 & \multicolumn{1}{l|}{\textbf{0.308}} & 0.038 & -0.039 \\
\multicolumn{1}{l|}{BERTScore R\_art} & 0.360 & 0.365 & 0.141 & 0.153 & 0.112 & \multicolumn{1}{l|}{0.234} & 0.144 & -0.022 \\
\multicolumn{1}{l|}{BERTScore F1\_art} & 0.358 & 0.365 & 0.230 & 0.256 & 0.192 & \multicolumn{1}{l|}{0.307} & 0.111 & -0.040 \\ \hline
\multicolumn{9}{l}{Neural entailment} \\ \hline
\multicolumn{1}{l|}{FactCC} & 0.376 &  &  &  &  & \multicolumn{1}{l|}{} & 0.071 &  \\
\multicolumn{1}{l|}{Dep Entail} & 0.342 &  &  &  &  & \multicolumn{1}{l|}{} & 0.092 &  \\ \hline
\multicolumn{9}{l}{Q \& A based} \\ \hline
\multicolumn{1}{l|}{FEQA} & -0.008 &  &  &  &  & \multicolumn{1}{l|}{} & 0.006 &  \\
\multicolumn{1}{l|}{QAGS} & 0.206 & 0.274 &  &  &  & \multicolumn{1}{l|}{} & -0.006 & \textbf{0.153} \\
\multicolumn{1}{l|}{QAEval - F1} &  &  &  &  &  & \multicolumn{1}{l|}{0.220 *} & -0.006 & \textbf{0.153} \\ \hline
\multicolumn{9}{l}{Text Generation based} \\ \hline
\multicolumn{1}{l|}{CBMI ( BART\_base + cnn )} & 0.058 & 0.026 & 0.152 & -0.029 & 0.023 & \multicolumn{1}{l|}{0.208} & -0.077 & -0.041 \\
\multicolumn{1}{l|}{BARTScore ( BART\_large + cnn )} & 0.413 & 0.470 & 0.197 & 0.310 & 0.181 & \multicolumn{1}{l|}{0.263} & 0.137 & 0.072 \\
\multicolumn{1}{l|}{BARTScore ( BART\_large + cnn + para )} & 0.392 & 0.416 & 0.259 & 0.301 & 0.238 & \multicolumn{1}{l|}{0.278} & 0.145 & 0.031 \\ \hline
\multicolumn{9}{l}{Proposed} \\ \hline
\multicolumn{1}{l|}{HaRiM ( BART\_large + cnn )} & \textbf{0.424} & \textbf{0.478} & 0.251 & \textbf{0.315} & 0.210 & \multicolumn{1}{l|}{0.284} & 0.136 & 0.076 \\
\multicolumn{1}{l|}{HaRiM ( BART\_large + cnn + para )} & 0.399 & 0.401 & \textbf{0.281} & 0.293 & \textbf{0.245} & \multicolumn{1}{l|}{0.282} & 0.141 & 0.028 \\ \hline
\end{tabular}%
}
\end{table}

They perform experiments on various datasets: FRANK~\cite{pagnoni-etal-2021-understanding}, and
QAGS annotations~\cite{wang-etal-2020-asking}. FRANK and QAGS contain 2,246 and 470 pairs, respectively, of article and system-generated summary from CNN-DailyMail~\cite{nallapati2016abstractive} as well as BBC-XSUM~\cite{narayan-etal-2018-dont} corpora. Every example in the benchmark contains human judgement on the corresponding example's factuality.

Table~\ref{tab:harim_result} shows the metric to human judgement (segment-level) correlation. From the table, we can observe that Harim+ shows the highest Kendall's $\tau$ in most criteria of CNN/DailyMail based benchmarks. In addition, Harim+ has the most correlation w.r.t. human judgements except several settings, i.e. XSUM and SummEval-Relevance. Another remark to make is that the proposed Harim+ prefers self-generated summary (i.e. summary generated by the same summarization model the scorer depending on) to human written references. This is likely due to the inductive bias of preferences toward summaries generated by abstractive summarization systems, as the score is based on an LLM.  

\subsubsection[HaluEval: A Large-Scale Hallucination Evaluation Benchmark for Large Language Models]{HaluEval: A Large-Scale Hallucination Evaluation Benchmark for Large Language Models~\cite{HaluEval}}
\textbf{Github Link: } https://github.com/RUCAIBox/HaluEval
\newline
\textbf{Tasks: } Question answering, knowledge-grounded dialogue, and summarization.
\newline
\textbf{Core Idea: } This paper proposes a benchmark for evaluating hallucinations in NLG tasks. The main framework is based on ChatGPT in a sampling-then-filtering setup. 
\newline
\newline
In this work, the authors propose the Hallucination Evaluation for Large Language Models (HaluEval) benchmark, a large collection of generated and human-annotated hallucinated samples for evaluating the performance of LLMs in detecting hallucination. To generate these samples, the authors propose a ChatGPT-based two-steps framework, i.e., sampling-then-filtering. In addition, this work also utilizes human experts to annotate the hallucinations in ChatGPT responses. The authors show experimentally that ChatGPT is likely to fabricate unverifiable information and hallucinate in specific topics. Moreover, the experiments show that it is challenging to detect the hallucinations in the texts using existing LLMs. Nevertheless, the authors show that empirically, one can improve the hallucination recognition ability of the LLMs by providing external knowledge or adding reasoning steps. 

\begin{figure}[t]
    \centering
    \includegraphics[width=\linewidth]{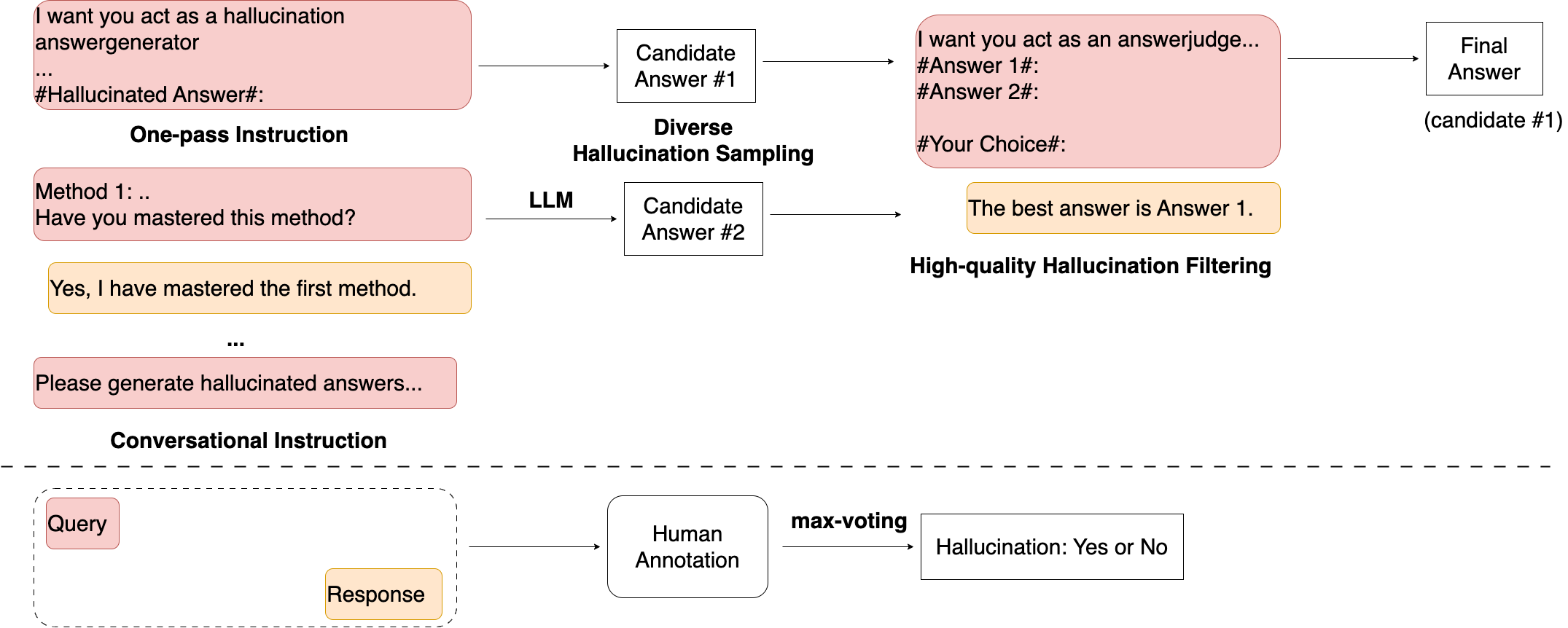}
    \caption{Creation pipeline of our benchmark, including automatic generation (top) and human annotation (bottom).~(Figure source:~\cite{HaluEval})}
    \label{fig:halueval_dataset}
\end{figure}

The synthetic dataset generation pipeline includes two steps: 1) diverse hallucination sampling, and 2) high-quality hallucination filtering. Both of these two steps relies heavily on prompt engineering. To sample diverse hallucinations, the authors propose two different hallucination sampling methods. For both of these methods, different instructions of hallucination sampling are proposed and followed by ChatGPT. As shown in Figure~\ref{fig:halueval_dataset}, the first method adopts a one-pass instruction process, where a hallucinated response is obtained through one feed of prompt. As the second method, the authors use a conversational instruction schema, where ChatGPT is taught to successively learn part of the instruction via a prompt engineering based approach. Then, based on the learned instructions, ChatGPT will generate another hallucinated answer, different from the first method, resulting in diverse and multi-facet hallucinated answers for each question. The next step in the generation pipeline is the so-called high-quality hallucination filtering. To construct a challenging benchmark for evaluation, the authors aim to select the most plausible and difficult hallucinated sample from the above two sampling methods. The authors design the prompt of the hallucination filtering to select the best answer from two candidates. In the instruction of filtering, the demonstration includes the ground-truth correct answer and a hallucinated counterpart. However, in the test example, the inputs are swapped into two hallucinated answers obtained from the previous step. Through this process, the most plausible hallucinated answer is expected to be selected by ChatGPT. This way, the final selected sample can be difficult to identify. 

The authors then ask human experts to annotate the general user queries and ChatGPT responses from the 52K instruction tuning dataset from Alpaca ~\cite{alpaca}. A pre-selection method is applied to obtain the most challenging examples. The authors use ChatGPT to sample three responses for each user query and compute their average semantic similarity using BERTScore~\cite{bert-score}. In the end 5, 000 user queries are retained with the lowest similarities. For each query and ChatGPT response, human experts will annotate whether the response contains hallucinated information ("Yes" or "No") and list the corresponding spans. 

Table~\ref{tab:halueval_result} presents results regarding the accuracy of several LLMs to recognize whether a sample (e.g., an answer, a dialogue response, or a summary) contains hallucinated information. One can observe that LLMs are typically not an expert at identifying hallucination. For example, ChatGPT cannot distinguish between factual and hallucinated summary and only achieves 58.53$\%$ accuracy in text summarization, which is barely above chance. 

\begin{table}[t]
\centering
\caption{Accuracy ($\%$) of evaluation models to classify whether the sample contains hallucinated contents in HaluEval benchmark.~(Table source:~\cite{HaluEval})}
\label{tab:halueval_result}
\resizebox{0.6\textwidth}{!}{%
\begin{tabular}{lllll}
\hline
Models & QA & Dialogue & Summa. & General \\ \hline
GPT-3 (davinci) & 49.21 & 20.02 & 51.23 & 77.54 \\
text-davinci-002 & 60.05 & 60.81 & 47.77 & 87.60 \\
text-davinci-003 & 49.65 & 68.37 & 48.07 & 87.54 \\
ChatGPT & 62.59 & 72.40 & 58.53 & 86.22 \\ \hline
\end{tabular}%
}
\end{table}

The authors also explore three improvement on the ability of LLMs to recognize hallucination: knowledge retrieval, chain of thought reasoning prompt and sample contrast. For knowledge retrieval, the authors provide the LLM with additional facts from Wikipedia to improve the detection results. For chain of though (CoT) reasoning prompt, the authors follows the idea of CoT from~\cite{wei2022chain} and construct the corresponding prompt to perform reasoning and derive the final answer by introducing a series of intermediate reasoning steps. For sample contrast, the authors further provide ground truth examples for ChatGPT to test whether it can distinguish the right sample from the hallucinated sample. As one can observe from the results in Table~\ref{tab:halueval_result_improve}, though limited, knowledge retrieval and CoT can offer certain degree of improvements over some tasks while sample contrast, on the contrary, is detrimental to the performance of the detection. 

\begin{table}[t]
\centering
\caption{Accuracy ($\%$) of ChatGPT equipped with three improvement strategies.~(Table source:~\cite{HaluEval})}
\label{tab:halueval_result_improve}
\resizebox{0.6\textwidth}{!}{%
\begin{tabular}{lllll}
\hline
Models & QA & Dialogue & Summa. & General \\ \hline
ChatGPT & 62.59 & 72.40 & 58.53 & 86.22 \\
w/Knowledge & 76.83 & 73.80 & - & - \\
w/CoT & 59.58 & 71.39 & 61.21 & 86.50 \\
w/Contrast & 40.19 & 68.67 & 49.46 & - \\ \hline
\end{tabular}%
}
\end{table}

\section{Hallucination Mitigation}
\label{sec: mitigation}
Hallucination mitigation is the task to reduce the potential hallucinations in the generated responses by LLMs. 
The papers reviewed in this section includes three categories: knowledge and retrieval-based approaches~\cite{ji2023rho, dziri2021neural, shuster2021retrieval}, which ground LLM responses in factual data using external knowledge sources such as knowledge graphs and retrieval systems. 
Another category is training or reference guiding for language models, involves strategies like employing control codes~\cite{rashkin2021increasing}, or contrastive learning~\cite{sun2023contrastive} to guide the generation process to discern between correct and hallucinated content. 
The third category is evaluation and mitigation on the language model generated content focusing on specific hallucination types, such as employing methods to evaluate quantity entity in summarization~\cite{zhao2020reducing}, and methods to detect and mitigate self-contradictory statements~\cite{mundler2023self}.

\subsection[RHO: Reducing Hallucination in Open-domain Dialogues with Knowledge Grounding]{RHO: Reducing Hallucination in Open-domain Dialogues with Knowledge Grounding~\cite{ji2023rho}}

\textbf{Github Link: } https://github.com/ziweiji/rho
\newline
\textbf{Task: } Knowledge-Grounded Dialogue
\newline
\textbf{Core Idea:} RHO is a conversational model that reduces hallucinations by merging knowledge graph triples with dialogue history and employing both local and global grounding techniques to ensure responses align with the provided knowledge and dialogue context.
\newline

RHO~\cite{ji2023rho} is a conversational model mitigating hallucination by leveraging local and global knowledge grounding techniques, combined with a response re-ranking mechanism, to ensure knowledge relevance in generated dialogue responses. The problem is defined as a special case of response generation task in dialogue systems: response generation for knowledge-grounded dialogue (KGD) task. KGD refers to a conversation where one or more than one of the participants uses external knowledge sources to drive the response generation. Response generation in dialogue systems involves using a dialogue history including a set of utterances from human participants. The KGD task uses a multi-relational knowledge graph (KG) as additional input in addition to the dialogue history. The goal is to generate a faithful response considering both the dialogue history and a subset of the relevant part of the KG.

The dataset used is OpenDialKG~\cite{moon2019opendialkg}, which contains open-ended dialogues between two speakers, initiated by talking about a given entity (real-world objects or concepts such as people, places, things, ideas, or events) and dialog turns are optionally paired with corresponding grounded Knowledge Graph (KG) paths of relevant facts.In addition to the dialogues, KG triples (subject, relation, object) are also provided in the dataset. A path in a KG refers to a sequence of triples that connect multiple entities through their relationships. The sequential dialog turns can be regarded as traversing the paths in the KG. The authors filter OpenDialKG by keeping only the dialogue samples that are annotated with a KG path. Statistically, OpenDialKG contains 13,802 dialog sessions containing 91,209 total turns about four domains: movie, book, sports, and music. The KG triples in the dataset are of 100,813 entities, 1,358 relations, 1,190,658 triples.

The RHO method starts with linearizing the related triples from the KG into a text format and merging it with the dialogue history. 
To have not only the lexical but also structured knowledge information, the authors do local knowledge grounding. 
%
%
%
TransE~\cite{bordes2013translating} is used to produce embeddings of entities and relation predicates from the entire KG. 
Then a locally grounded token embedding is produced for an arbitrary token in the dialogue history.
If the token is a substring of an entity, it is mapped to the entity's embedding; If the token is a substring of a relation, it is mapped to the relation's embedding.
The local embedding is produced to represent a token's association with entities or relations from a knowledge graph. Local knowledge grounding connect tokens from a dialogue to relevant information within a KG.
In addition to local grounding, the authors further propose global knowledge grounding to draw global dependencies between the dialogue history and the representations of all triples (subject, predicate, and object) in the context-related sub-graph.
For each triple in the sub-graph, an embedding vector is created by combining the mapped embeddings of the subject, predicate, and object of the triple.
%
%
%
Then global knowledge embedding space is formed by projecting and concatenating the embedding vectors of all the triples in the sub-graph.
For each token in the dialogue linked to any entity or relation, the global token embedding is calculated by applying a softmax function to the product of the token's embedding and transposed global knowledge embedding space. This global embedding is produced based on the association with entities or relations.
The encoder then integrates both local and global embeddings for each token during training.
The training process involves an attention mechanism with respect to the entire sub-graph to determine how much each token should focus on different parts of global embedding space.
The above encode and decoder generate $N$ candidate responses.
%
%
A conversational reasoning model is then used to simulate actions that represent walking steps on the knowledge graph, These actions are based on the dialogue history and each candidate response.
The actions derived from walking on the knowledge graph are constructed using embeddings of entities and relations. The mechanism calculates the probability of the actions aligning with the graph and the dialogue history for each candidate response. The response with the highest probability is then selected as the optimal.
%


\begin{table}[t]
\centering
\caption{Results from the automatic evaluation of RHO and its baselines are presented, where ``RD'' refers to reference-dependent mode, ``R'' refers to reference-free mode, and ``Pre'' refers to Precision. The results from the ablation study are presented in the last four rows. ``LKG'', ``GKG'', and ``RR'' correspond to local knowledge grounding, global knowledge grounding, and response re-ranking, respectively. ``Full Implementation'' encompasses all three elements combined, i.e., LKG$+$GKG$+$RR.~(Table source:~\cite{ji2023rho})} 
\resizebox{0.8\textwidth}{!}{
\begin{tabular}{@{}lllllllll@{}}
\toprule
Model & \multirow{2}{*}{BLEU4} & \multirow{2}{*}{ROUGE - L} & \multirow{2}{*}{FeQA} & \multicolumn{2}{l}{QuestEval} & \multicolumn{3}{l}{Entity Coverage (\%)} \\
 &  &  &  & RD & RF & Pre & Recall & F1 \\ \midrule
EARL & 7.97 & 23.61 & 39.93 & 37.88 & 35.59 & 86.61 & 45.17 & 64.44 \\
GPT2 & 10.27 & 29.59 & 39.60 / 26.54 & 46.86 & 42.07 & 91.62 & 33.26 & 52.30 \\
GPT2 + NPH & 10.41 & 29.93 & 40.83 / 28.98 & 47.45 & 42.45 & 95.61 & 33.39 & 53.96 \\
BART & 14.45 & 33.33 & 39.00 & 46.97 & 42.75 & 96.99 & 44.96 & 62.87 \\
BART + NPH & 15.53 & 34.99 & 42.41 & 47.94 & 43.56 & 96.44 & 44.12 & 65.98 \\
KG - BART & 13.72 & 33.31 & 41.87 & 45.55 & 42.86 & 97.68 & 45.63 & 64.58 \\
RHO ( LKG ) & 19.89 & \textbf{39.95} & 43.04 & 48.91 & 44.37 & 97.38 & 45.57 & 67.77 \\
RHO ( GKG ) & \textbf{20.77} & 39.54 & 40.65 & 48.41 & 43.84 & 97.20 & 45.63 & 67.40 \\
RHO ( LKG + GKG ) & 20.63 & 39.51 & 45.96 & 50.35 & 46.03 & 98.26 & 50.74 & 71.47 \\
RHO ( Full Implementation ) & 19.11 & 38.45 & \textbf{47.99} & \textbf{50.58} & \textbf{46.41} & \textbf{98.53} & \textbf{51.77} & \textbf{72.29} \\ \bottomrule
\end{tabular}}
\end{table}

In the evaluation conducted , BART and GPT2+NPH are noteworthy baselines. BART achieved a BLEU4 score of 14.45 and an F1 score of 62.87, while GPT2+NPH registered a BLEU4 score of 10.41 and an F1 score of 53.96.
RHO implemented with both local and global knowledge grounding as well as response re-ranking, demonstrates superior performance across multiple evaluation metrics. Notably, RHO (GKG) achieves a BLEU4 score of 20.77, RHO (LKG) achieves a ROUGE-L of 39.95. Furthermore, the RHO (Full Implementation) configuration exhibits the highest Entity Coverage at 98.53, a FeQA score of 47.99 and an F1 score of 72.29. These results highlight the effectiveness of RHO, especially when all its components are implemented together.

\subsection[Neural Path Hunter: Reducing Hallucination in Dialogue Systems via Path Grounding]{Neural Path Hunter: Reducing Hallucination in Dialogue Systems via Path Grounding~\cite{dziri2021neural}}
\label{sec:nph}

\textbf{Github Link: } https://github.com/nouhadziri/neural-path-hunter
\newline
\textbf{Task: } Knowledge-Grounded Dialogue
\newline
\textbf{Core Idea:} 
The Neural Path Hunter (NPH) model reduces hallucinations in knowledge-grounded dialogue by employing a generate-then-refine strategy, where after a response was generated by an LLM, a token-level fact critic identifies potentially hallucinated entities which will be refined by querying and with an external knowledge graph.
\newline

\begin{figure}[!htbp]
    \begin{minipage}{0.60\textwidth}
        \centering
        \includegraphics[width=\linewidth]{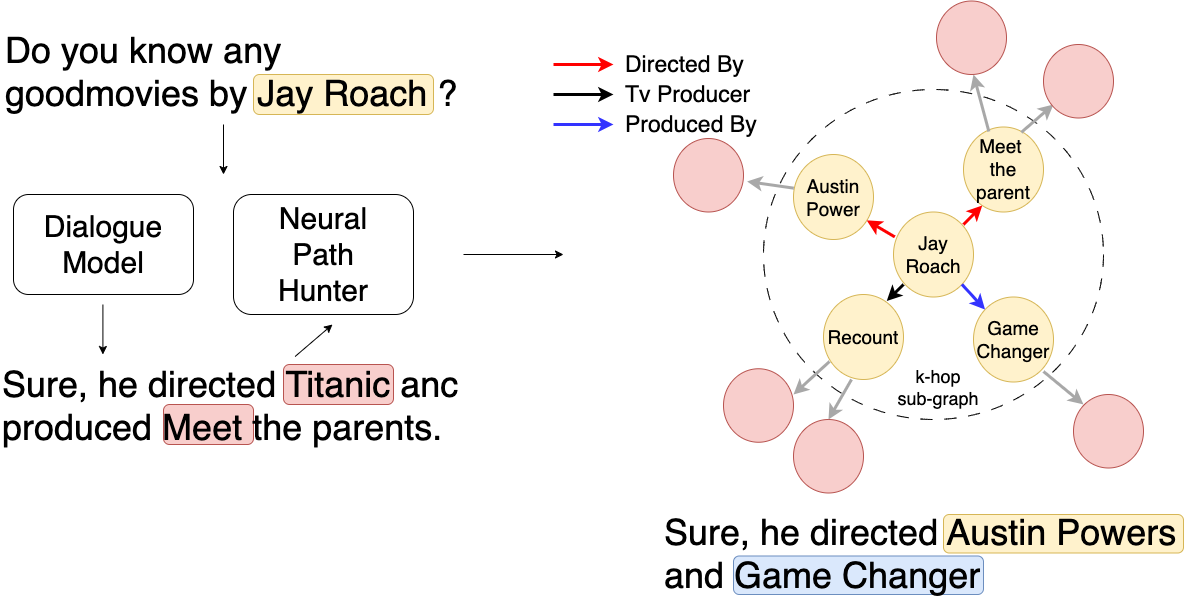}
        \label{fig:fig1}
    \end{minipage}%
    \hfill 
    \begin{minipage}{0.38\textwidth}
        \centering
        \includegraphics[width=\linewidth]{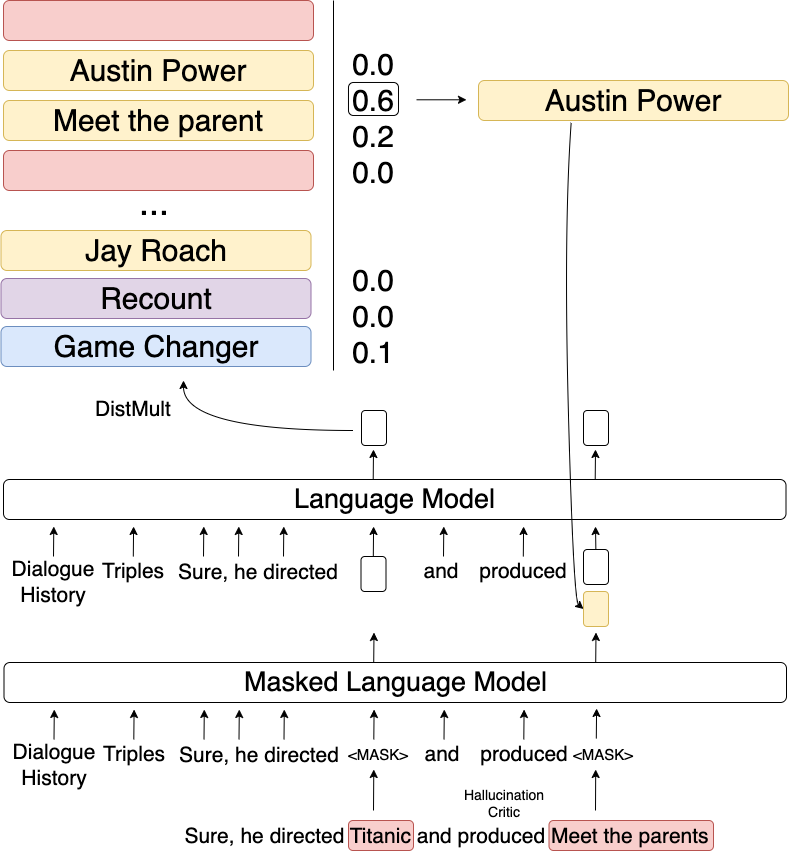}
        \label{fig:fig2}
    \end{minipage}
    
    \begin{minipage}{1\textwidth} 
        \centering
        \resizebox{0.75\textwidth}{!}{ 
        \begin{tabular}{@{}llll@{}} 
        \toprule
        Model & FeQA & Critic & BLEU \\ 
        \midrule
        GPT2 - KG & 26.54 & 19.04 & 11.79 * \\
        GPT2 - KG + NPH & 28.98 * & 11.72 * & 11.29 \\
        GPT2 - KG + NPH - W / O NCE & 26.02 & 17.91 & 10.98 \\
        GPT2 - KG + NPH - W . COMPGCN \qquad\qquad\qquad\qquad\qquad\qquad\qquad\qquad\qquad\qquad\qquad & 26.89 & 15.41 & 11.10 \\
        GPT2 - KG + NPH - W / O MLM & 27.01 & 15.02 & 10.88 \\
        GPT2 - KG + NPH - W / O CRITIC & 18.23 & 19.65 & 6.49 \\
        AdapterBot & 23.11 & 26.68 & 10.56 \\
        AdapterBot + NPH & 27.21 * & 18.51 * & 10.74 * \\
        AdapterBot + NPH - W / O NCE & 24.02 & 25.02 & 9.98 \\
        AdapterBot + NPH - W . COMPGCN & 25.83 & 20.23 & 10.11 \\
        AdapterBot + NPH - W / O MLM & 26.02 & 21.04 & 10.06 \\
        AdapterBot + NPH - W / O CRITIC & 16.21 & 27.22 & 5.64 \\
        GPT2 - KE & 19.54 & 28.87 & 6.24 * \\
        GPT2 - KE + NPH & 26.21 * & 20.34 * & 6.06 \\
        GPT2 - KE + NPH - W / O NCE & 20.34 & 24.32 & 5.89 \\
        GPT2 - KE + NPH - W . COMPGCN & 23.23 & 21.21 & 6.01 \\
        GPT2 - KE + NPH - W / O MLM & 24.01 & 22.40 & 5.99 \\
        GPT2 - KE + NPH - W / O CRITIC & 15.89 & 30.71 & 3.49 \\
        Gold response & 33.34 & 5.2 & - \\ 
        \bottomrule
        \end{tabular}}
        \label{fig:fig1x}
    \end{minipage}%
    \caption{(Top left): NPH overview. (Top right): Entity Mention Retriever architecture. (Bottom): Assessing the degree of hallucination of different models before and after refinement on the generated samples based on the OpenDialkg test data. A higher FeQA score suggests greater faithfulness. The hallucination Critic (Critic) measures the percentage of hallucinated responses in the dataset. (* p-value < 0.001). NPH employs GPT2 embeddings for the KG-Entity Memory.~(Figure and table source:~\cite{dziri2021neural})}
\end{figure}

The Neural Path Hunter (NPH)~\cite{dziri2021neural} employs a generate-then-refine approach for LLM-based dialogue systems. The problem is that knowledge-grounded dialogue systems often suffer from the hallucination problem. The dialogues generated were incorrect or fabricated. The problem could stem from the misuse of entities in the dialogue, leading to factual inaccuracies.

In NPH, a token-level fact critic was trained to identify potential hallucination entities in a sentence, focusing especially on entity misuse. The critic will be used in the pipeline of NPH to flag entities of concern in a sentence where a binary label is predicted at each word position. The NPH critic model leverages Roberta-Large from Huggingface: \footnote{ https://huggingface.co/roberta-large} for token classification with the training data manually introduced some negatives, i.e. replacing some correct entities to incorrect entities, or swapping the subject and object for the dialogues from OpenDialKG dataset.

The NPH pipeline starts with a language model (LM), such as GPT2, to generate a dialogue response. The critic will be used to flag potentially hallucinated entities. For each flagged entity, a representation is formed using a Masked Language Model (MLM), obtaining contextual representations, which are the representations of the sentence without that potentially hallucinates words (entities). These representations undergo a pooling operation to attain a singular representation for each entity and then an auto-regressive LM then formulates a query from the entity representation. The system consults the Knowledge Graph (KG) using this query, substantiating the entities by navigating the KG. KG representation is standardized using KG-Entity Memory with two methods: GPT2 embeddings or CompGCN, a Graph Convolutional Network designed for multi-relational data. A scoring function DistMult~\cite{wang2014knowledge}, evaluates each KG-Entity memory triple, selecting the entity with the highest score as the factual entity. The original response is refined using this accurate entity, ensuring the dialogue's factual correctness, finishes the whole generate-then-refine process. By integrating the KG as an external oracle and utilizing the generate-then-refine pipeline, NPH enhances factual accuracy of the dialogue system.

The results show that using the NPH approach on the OpenDialkg test data, the FeQA score for GPT2-KG increased from 26.54 to 28.98, for AdapterBot from 23.11 to 27.21, and for GPT2-KE from 19.54 to 26.21. Concurrently, hallucination critic (the percentage of data that is predicted as potentially hallucinated by the pretrained critic hallucination detector) scores decreased, with GPT2-KG reducing from 19.04 to 11.72, AdapterBot from 26.68 to 18.51, and GPT2-KE from 28.87 to 20.34. 

\subsection[Retrieval Augmentation Reduces Hallucination in Conversation]{Retrieval Augmentation Reduces Hallucination in Conversation~\cite{shuster2021retrieval}}

\textbf{Github Link: } https://parl.ai/projects/hallucination/
\newline
\textbf{Task: } Knowledge-Grounded Dialogue
\newline
\textbf{Core Idea:} The Retrieval-Augmented Generation (RAG)-based method proposed reduces hallucination in knowledge-grounded dialogue systems by searching, ranking, and incorporating external documents into the encoder-decoder process, using mechanisms like Dense Passage Retrieval~\cite{karpukhin2020dense} and Poly-encoders~\cite{humeau2019poly}, to produce factually-grounded responses.
\newline

The paper discusses the knowledge-grounded dialogue task. Studies indicated that Retrieval-Augmented Generation (RAG) methods have been effective for producing more accurate responses in open-domain QA but have limitations when applied directly to multi-turn dialogue contexts. The challenge associated is to explore the various techniques related to RAG to be able to adaptively apply RAG to open-domain knowledge-grounded dialogue. The dataset used is Wizard of Wikipedia, the same as in Section~\ref{control_code}.

RAG employs an encoder-decoder to encode questions and generate (decode) answers. The encoding process is enhanced with documents or passages retrieved from a set of documents using a learned matching function. The authors studied various types of architectures with multiple components of a broader category of RAG models: retrieval-augmented neural architectures. The components include retrievers, rankers, and encoder-decoders.When a user has a prompt or a query. The retriever component takes this prompt and searches for relevant documents from a large corpus. Once a set of potentially relevant documents is retrieved, the ranker sorts these documents based on their relevance to the user's prompt. This step ensures that the most pertinent information is prioritized. The encoder processes the user's prompt along with the top-ranked documents retrieved and produces an intermediate representation. This representation captures the essential information from both the user's input and the external knowledge sourced from the database. The decoder then uses this representation to generate a response. The encoder-decoder enables the system to provide responses that are grounded in factual information.

The authors discussed various retrieval augmentation mechanisms.Dense Passage Retrieval (DPR)~\cite{karpukhin2020dense} is a retrieval strategy that employs a dual-encoder architecture to score dialogue context-document pairs. Given a dialogue context, DPR retrieves relevant documents by computing a relevance score between the context and each document in a corpus. It encodes the queries and the documents separately. The term dense opposes sparse retrieval method, as it maps both the query and the documents into continuous vector spaces. Poly-encoders~\cite{humeau2019poly} produce a set of vectors (codes) to attentively process the context token outputs of a transformer encoder. These codes capture different distinct semantic aspects or features of the context. Each code interacts with a candidate's vector to compute a score. These scores from all codes are then combined for a final score for that candidate.The Fusion-in-Decoder (FiD)~\cite{izacard2021leveraging} mechanism fuses independent encoder outputs before decoding the final generation. While DPR and Poly-encoder are retrieval strategies used to find relevant documents or passages given the dialogue context, FiD is for integrating the information from these retrieved documents into the final response generated by the model. FiD then separately encodes the dialogue context and each retrieved document using a transformer-based encoder. The separate encodings are then fused together (combining all of the outputs from the encoder) in the decoding step.


\begin{table}[t]
\centering
\caption{Comparison of Seq2Seq Models and Retrieval Augmentations on Wow Test (Seen). Perplexity (PPL) scores aren't directly comparable between various seq2seq models due to their use of distinct dictionaries. Retrieval models are retrieving five documents over all of Wikipedia. All RAG models are RAG-Token.~(Table source:~\cite{shuster2021retrieval})} 
\resizebox{0.8\textwidth}{!}{
\begin{tabular}{@{}lllllll@{}}
\toprule
Seq2Seq Model & Retrieval Mechanism & PPL & F1 & Knowledge F1 & BLEU-4 & ROUGE-L \\ \midrule
BlenderBot - 400m & None & 11.2 & 19.7 & 16.3 & 1.4 & 18.8 \\
 & RAG DPR & 9.0 & 21.1 & 23.7 & 3.0 & 21.2 \\
 & RAG DPR - Poly & 9.7 & 21.1 & 24.2 & 3.0 & 21.0 \\
BART - Large & None & 14.7 & 20.9 & 17.4 & 1.7 & 20.3 \\
 & FiD & 13.7 & 20.8 & 21.5 & 2.5 & 21.2 \\
 & RAG DPR & 12.7 & 22.4 & 22.5 & 3.4 & 22.9 \\
 & RAG DPR - Poly & 11.4 & \textbf{22.9} & 26.5 & 3.9 & \textbf{23.5} \\
 & FID - RAG DPR & 11.8 & 21.1 & 29.6 & 3.8 & 22.7 \\
 & FID - RAG DPR - Poly & 11.4 & 22.1 & \textbf{29.7} & \textbf{4.1} & 23.0 \\
T5 Large & None & 12.1 & 19.3 & 14.6 & 1.0 & 18.1 \\
 & RAG DPR & 9.8 & 21.9 & 25.9 & 3.8 & 22.1 \\
 & FID - RAG DPR & 9.5 & 22.0 & 27.8 & 3.9 & 22.3 \\ \bottomrule
\end{tabular}
}
\label{tab:RAG_0} 
\end{table}

Table~\ref{tab:RAG_0} demonstrates the performance of retrieval strategies on the Wizard of Wikipedia Test (Seen) dataset. The RAG DPR-Poly version achieves a Knowledge F1 score of 26.5 and a ROUGE-L rating of 23.5, respectively. Additionally, FiD-RAG DPR-Poly demonstrates improved results, securing a Knowledge F1 score of 29.7 and a ROUGE-L score of 23.0. For the BLEU-4 metric, FiD-RAG DPR-Poly achieves a score of 4.1, notably higher than BART-Large None score of 1.7.

\subsection[Increasing Faithfulness in Knowledge-Grounded Dialogue with Controllable Features]{Increasing Faithfulness in Knowledge-Grounded Dialogue with Controllable Features~\cite{rashkin2021increasing}}
\label{control_code}

\textbf{Github Link: } N/A
\newline
\textbf{Task: } Knowledge-Grounded Dialogue
\newline
\textbf{Core Idea:} The control codes introduced reduce hallucination in knowledge-grounded dialogue systems by having these special tokens named control codes to guide the language model in producing responses that emphasize lexical precision, objective voice, and entailment; this is achieved by integrating these codes during the training phase to reduce hallucinations and by employing these code in a resampling method during the decoding phase.
\newline

Rashkin et al.\ proposed a method to enhance the faithfulness of responses in knowledge-grounded dialogue systems by incorporating control codes~\cite{rashkin2021increasing}. The problem is how to ensure that make knowledge-grounded dialogue systems generated conversational responses remain faithful to the provided evidence, especially when the conversation datasets contain both evidence-based and subjective responses. The authors introduced control codes, which are special tokens that will be used to influence the LM by being added to the input or designed to guide the output from language models.

\begin{figure}[t]
    \centering 
    \includegraphics[width=0.9\linewidth]{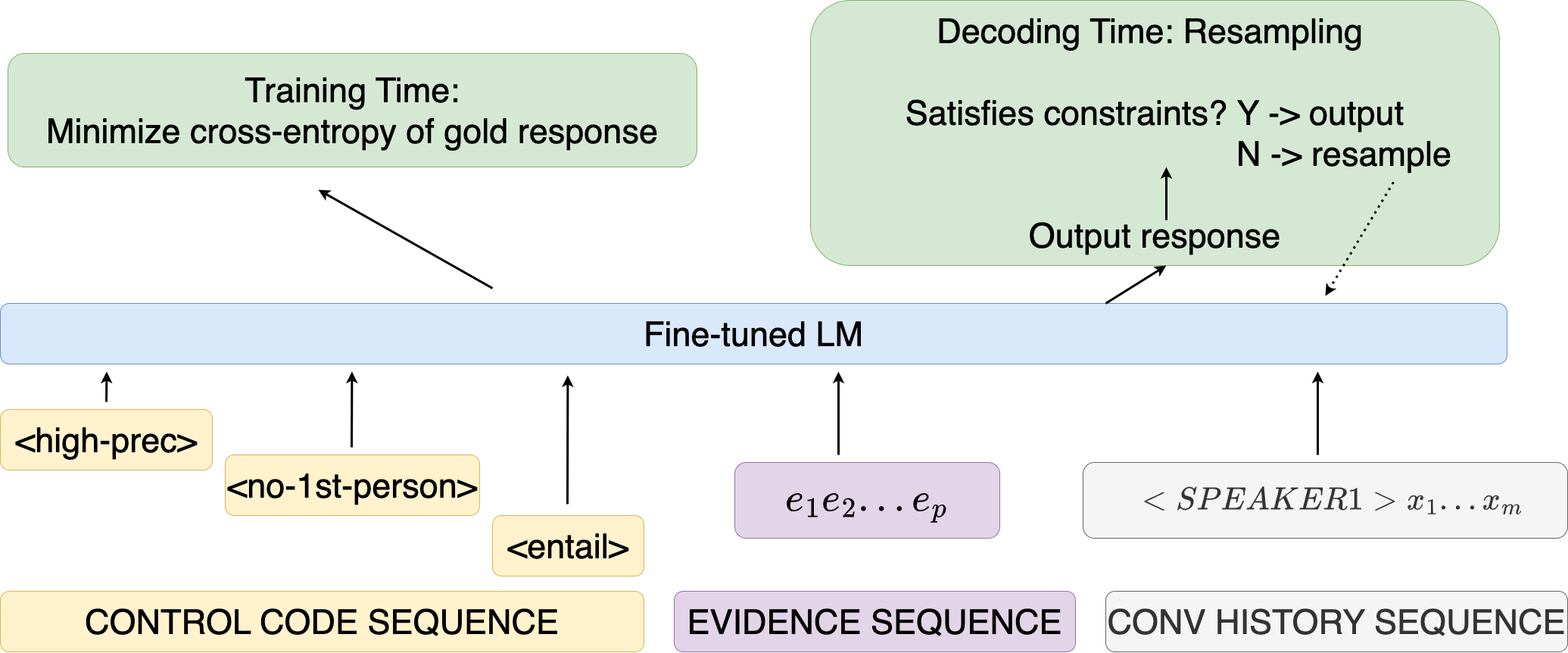} 
    \caption{System framework: a large pre-trained language model is employed to encode both the evidence and conversation history leading to produce a response. Additional tokens (i.e. control codes) are incorporated to train the model to recognize differences between types of utterance based on their alignment with the evidence. During the decoding phase, the authors investigate the significance of using resampling methods.~(Figure source:~\cite{rashkin2021increasing})}
    \label{fig:CC_0} 
\end{figure}

To produce more grounded responses that adhere closely to source documents, the authors proposed two control code-based methods: one that integrates control code features based on evaluation measures as special tokens prepended to the dialogue inputs of during training. The combination of the control codes is to reduce the hallucination by forcing the model to be more aware of how the response relies on the knowledge evidence in the decoding to avoid speculation. The other method employs a form of resampling to directly restrict the output to match the proposed evaluation measures.  During the decoding, the resampling method is used to continue to sample until found one satisfied the evaluation measures. The evaluation measures mentioned in both methods include objective voice, lexical precision, and entailment. Lexical precision is calculated as the proportion of words in the generated response that are present in the evidence.  Objective voice is assessed by checking the presence of first-person pronouns in the response. This is a binary measure,  where a response is marked as objective (having an objective voice). Entailment is a measure for to which extent the response is semantically entailed by the source document. It is calculated using a pretrained Natural Language Inference (NLI) model.

\begin{table}[t]
\centering
\caption{Experimental results on the seen/unseen topic portions of the Wizard of Wikipedia test set. The metrics presented include BLEU-4, the percentage of replies that avoid using the first person (N1P), precision and recall of words in the response with the evidence, and the fraction of responses that are predicted to be entailed by the evidence.~(Table source:~\cite{rashkin2021increasing})} 
\resizebox{0.8\textwidth}{!}{
\begin{tabular}{@{}lcccccccccc@{}}
\toprule
 & \multicolumn{5}{c}{Seen Topic} & \multicolumn{5}{c}{Unseen Topic} \\ \midrule
\multirow{2}{*}{Model} & BLEU & N1P & \multicolumn{2}{c}{w / Evid.} & NLI & BLEU & N1P & \multicolumn{2}{c}{w / Evid.} & NLI \\
 & B4 & \% & Prec & Rec & \% & B4 & \% & Prec & Rec & \% \\ \midrule
E2E model & 1.5 & 48.0 & 47.9 & 30.4 & 29.3 & 0.3 & 37.6 & 33.2 & 21.7 & 9.5 \\
dodecaDialogue & \textbf{10.0} & 78.3 & 81.1 & \textbf{67.7} & 70.3 & \textbf{9.7} & 77.7 & 81.3 & \textbf{66.5} & 70.6 \\
GPT - 2 ( none ) & 6.2 & 50.9 & 56.1 & 49.4 & 34.2 & 5.7 & 52.1 & 56.4 & 48.3 & 34.2 \\
GPT - 2 ( control codes ) & 7.8 & 99.3 & 76.6 & 61.5 & 73.8 & 7.6 & 99.2 & 77.0 & 60.3 & 74.0 \\
GPT - 2 ( resampling ) & 7.6 & 75.1 & 70.4 & 57.7 & 71.4 & 7.2 & 76.2 & 70.3 & 56.5 & 72.3 \\
GPT - 2 ( both ) & 8.9 & \textbf{99.9} & 83.1 & 66.3 & 93.9 & 8.4 & \textbf{99.8} & 83.2 & 64.7 & \textbf{94.4} \\
T5 ( none ) & 7.6 & 51.1 & 64.0 & 51.9 & 45.1 & 7.4 & 51.4 & 65.2 & 51.9 & 44.9 \\
T5 ( control codes ) & 8.6 & 99.7 & 84.3 & 62.1 & 89.0 & 8.5 & 99.4 & 85.0 & 61.5 & 89.8 \\
T5 ( resampling ) & 8.2 & 77.5 & 73.3 & 55.5 & 74.7 & 8.1 & 78.5 & 74.4 & 55.5 & 76.3 \\
T5 ( both ) & 8.4 & 99.8 & \textbf{85.0} & 62.1 & \textbf{94.0} & 8.7 & \textbf{99.8} & \textbf{86.1} & 62.2 & \textbf{94.4} \\ \bottomrule
\end{tabular}
}
\label{fig:CC_1} 
\end{table}

The dataset used is th Wizard of Wikipedia~\cite{dinan2018wizard}, a multi-turn knowledge-grounded dialogues dataset. The conversation is between an apprentice and a wizard. The wizard has access to Wikipedia documents. Evidence spans from the documents are labeled for every utterance by the wizard. The dataset consists of 73,571 training responses, with development sets of 3,905 for seen topics and 3,898 for unseen topics, and test sets of 3,842 for seen topics and 3,902 for unseen topics. 
It is noteworthy that 44\% of the training responses employ first-person language, with an average lexical precision with respect to evidence at 0.43, and only 23\% of the responses being predicted as entailed by the evidence.
%


%
The performance of various models on the Wizard of Wikipedia test sets of seen topic and unseen topic in the Table~\ref{fig:CC_1}. The GPT-2 model, when enhanced with control codes, achieved a BLEU-4 score of 7.8, 7.6, up from its baseline of 6.2, 5.7. When integrated with resampling and control code, the model achieves the highest NIP of 99.9, 99.8 on seen topic and unseen topic, respectively. Similarly, the T5 model demonstrated improvements, achieving an NIP of 99.8 , NLI of 94.4 on unseen topic.

\subsection[Contrastive Learning Reduces Hallucination in Conversations]{Contrastive Learning Reduces Hallucination in Conversations~\cite{sun2023contrastive}}

\textbf{Github Link: } https://github.com/sunnweiwei/mixcl
\newline
\textbf{Task: } Knowledge-Grounded Dialogue
\newline
\textbf{Core Idea:} The Mixed Contrastive Learning (MixCL) method reduces the hallucination produced by pre-trained language models in conversational systems by applying contrastive learning and data mixing at a fine-grained span level, emphasizing span-level contrasts to differentiate between correct and hallucinated entities without full retraining.
\newline

Pre-trained language models used in conversational systems may hallucinate by generating plausible but factually incorrect statements. The challenge is that the commonly used training method, maximum likelihood estimation (MLE) with teacher forcing, encourages models to merely replicate training data, leading to inaccuracies during querying real-world information. In order to tackle this issue, the authors proposes Mixed Contrastive Learning (MixCL). This method combines contrastive learning with data mixing to help the model differentiate between correct and hallucinated information. Firstly, through negative sampling, MixCL identifies confusing information (negative knowledge) that the model is likely to mix up. With mixed contrastive learning, the method contrasts this negative knowledge with the correct positive knowledge to create a mixed training example. MixCL operates at a fine-grained span level, using named entity recognition (NER) for intrinsic hallucinations and constituency parsing for extrinsic ones, to ensure precision in the knowledge it presents. 
This approach helps in training the LM to be more accurate in its responses.
The dataset is the Wizard-of-Wikipedia the same as described in section \ref{control_code}. 

\begin{figure}[t]
    \centering 
    \includegraphics[width=1.0\linewidth]{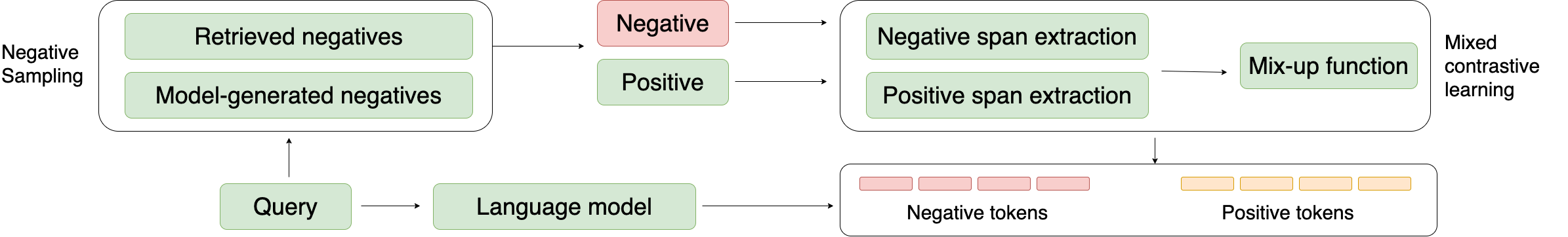} 
    \caption{Overview of MixCL: MixCL is comprised of two steps. The first step, (i) negative sampling: generates negative knowledge to produce the data for the model. The second stage, (ii) mixed contrastive learning, aims to reduce the generation of negative tokens by employing mixed contrastive learning.~(Figure source:~\cite{sun2023contrastive})} 
    \label{fig:MixCL_0} 
\end{figure}

Figure~\ref{fig:MixCL_0} presents an overview of the MixCL method. The method consists of two primary steps: negative sampling and mixed contrastive learning. The negative sampling step produces the most confusing negative knowledge samples that the model might misinterpret. Two techniques are employed for negative sampling: retrieved negatives, which uses a retrieval tool to obtain irrelevant knowledge from the corpus, and model-generated negatives, which exploits a model bootstrapping approach to produce knowledge where hallucinations exist. The mixed contrastive learning step emphasizes fine-grained span-level contrasts by extracting meaningful spans from both positive and negative knowledge, mixing these spans to produce new examples, and then optimizing the model through contrastive learning with a mixed-contrast loss function at the span level. 
%
%
In the MixCL method, the loss function is MixCL is optimized with a combination of three loss functions: $ J(\theta) = \alpha_1 L_{\text{MLE}} + \alpha_2 L_{\text{MCL}} + \alpha_3 L_{\text{LM}}$ (\( \alpha_1, \alpha_2, \alpha_3 \) are weights for the corresponding losses). The mixed-contrast loss ($L_{\text{MCL}}$) function is to train the model's ability to discern between reliable and hallucinated content. Complementing this, the Maximum Likelihood Estimation ($L_{\text{MLE}}$) loss function promotes the replication of training data patterns for the model's general language generation capabilities. Additionally, the language modeling loss function ($L_{\text{LM}}$) is employed to alleviate knowledge forgetting.
%


\begin{table}[t]
\centering
\caption{Evaluation results from the Wizard-of-Wikipedia dataset. The first group lists KB-based methods under realistic conditions. The second group lists KB-based methods under oracle conditions. The third group lists LM-based methods, including MixCL. Notably, MixCL results that markedly surpass the  previous-best LM-based approaches. And MixCL outcomes the leading KB-based methods under realistic conditions.~(Table source:~\cite{sun2023contrastive})} 
\resizebox{0.999\textwidth}{!}{
\begin{tabular}{@{}lllllllllllllllll@{}}
\toprule
 & \multicolumn{8}{c}{Test seen} & \multicolumn{8}{c}{Test unseen} \\ \midrule
Method & \multicolumn{1}{c}{F1} & \multicolumn{1}{c}{RL} & \multicolumn{1}{c}{B2} & \multicolumn{1}{c}{B4} & \multicolumn{1}{c}{MT} & \multicolumn{1}{c}{KF1} & \multicolumn{1}{c}{EF1} & \multicolumn{1}{c}{Acc} & \multicolumn{1}{c}{F1} & \multicolumn{1}{c}{RL} & \multicolumn{1}{c}{B2} & \multicolumn{1}{c}{B4} & \multicolumn{1}{c}{MT} & \multicolumn{1}{c}{KF1} & \multicolumn{1}{c}{EF1} & \multicolumn{1}{c}{Acc} \\ \midrule
\textit{KB-based methods under realistic conditions} &  &  &  &  &  &  &  &  &  &  &  &  &  &  &  &  \\
TMN & 17.3 & 17.0 & 5.7 & 1.1 & 14.8 & 15.8 & 8.7 & 15.2 & 14.4 & 14.5 & 3.3 & 0.3 & 11.5 & 9.4 & 2.1 & 8.6 \\
DukeNet & 18.5 & 17.7 & 6.4 & 1.9 & 16.0 & 18.5 & 12.0 & 20.6 & 15.9 & 15.9 & 4.8 & 1.1 & 13.7 & 14.7 & 8.0 & 14.3 \\
KnowledGPT & 21.1 & 20.1 & 8.9 & 3.4 & 20.0 & 22.2 & 15.5 & 24.3 & 19.5 & 18.4 & 8.0 & 2.6 & 18.3 & 20.0 & 11.7 & 20.2 \\
KnowBART & 21.1 & 18.9 & 8.5 & 3.3 & 17.8 & 21.3 & 16.2 & 24.2 & 21.0 & 18.3 & 8.9 & 3.6 & 17.9 & 22.5 & 16.2 & 24.0 \\ \midrule
\textit{KB-based methods under oracle conditions} &  &  &  &  &  &  &  &  &  &  &  &  &  &  &  &  \\
DukeNet & 19.3 & 18.7 & 7.5 & 2.5 & 17.2 & 19.6 & 13.2 & 22.1 & 17.1 & 17.0 & 6.0 & 1.7 & 15.2 & 16.5 & 9.2 & 16.8 \\
KnowledGPT & 22.0 & 20.8 & 9.9 & 3.7 & 20.9 & 23.8 & 16.9 & 26.3 & 20.5 & 19.5 & 8.7 & 3.0 & 19.3 & 22.1 & 13.3 & 22.6 \\
KnowBART & 22.1 & 19.6 & 9.1 & 3.7 & 18.1 & 23.1 & 18.0 & 26.8 & 22.7 & 20.1 & 9.8 & 4.3 & 18.7 & 24.1 & 18.4 & 27.5 \\ \midrule
\textit{LM-based methods} &  &  &  &  &  &  &  &  &  &  &  &  &  &  &  &  \\
GPT-2 & 19.6 & 18.5 & 7.8 & 1.4 & 17.8 & 17.9 & 13.3 & 15.4 & 18.3 & 17.3 & 6.5 & 0.8 & 16.1 & 14.6 & 7.2 & 8.4 \\
BlenderBot & 18.8 & 19.4 & 7.7 & 2.3 & 18.0 & 18.2 & 13.1 & 16.7 & 17.8 & 16.9 & 5.5 & 0.8 & 15.0 & 15.7 & 7.1 & 9.6 \\
KnowExpert & 18.7 & 18.6 & 6.7 & 1.3 & 16.5 & 14.1 & 9.8 & 12.6 & 16.7 & 17.2 & 5.4 & 0.6 & 14.5 & 11.8 & 5.5 & 9.2 \\
MSDP & 17.8 & 16.5 & 6.1 & 1.9 & 18.2 & 21.7 & 13.9 & 18.4 & 16.9 & 16.1 & 5.5 & 1.1 & 16.2 & 20.3 & 8.4 & 16.1 \\ \midrule
MixCL & 21.6 & 20.5 & 9.2 & 2.7 & 20.5 & 22.3 & 16.3 & 20.4 & 19.6 & 18.8 & 7.4 & 1.4 & 18.0 & 18.0 & 11.6 & 14.4 \\ \bottomrule
\end{tabular}
}
\label{tab:MixCL_1} 
\end{table}

Table~\ref{tab:MixCL_1} presents evaluation results for various conversational models on the Wizard-of-Wikipedia dataset, categorized into KB-based methods under realistic conditions (retrieving knowledge from the entire Wikipedia corpus without narrowed down search space), KB-based methods under oracle conditions (models are provided with a smaller subset of the Wikipedia corpus that definitely contains the required knowledge), and LM-based methods. 
For the KB-based models under realistic conditions, KnowledGPT achieved the F1 score of 21.1. However, in the LM-based category, MixCL exhibits superior performance with an F1 score of 21.6 for the test seen set surpassing the KB-based models under realistic conditions. 
For test seen set, MixCL achieves a ROUGE-L of 20.5 and BLEU-2 of 9.2, displaying a notable advantage over other LM-based approaches. 

\subsection[Reducing Quantity Hallucinations in Abstractive Summarization]{Reducing Quantity Hallucinations in Abstractive Summarization~\cite{zhao2020reducing}}

\textbf{Github Link: } N/A
\newline
\textbf{Task: } Summarization
\newline
\textbf{Core Idea:} The HERMAN model proposed in this paper corrects the hallucinated quantitative entities in abstractive summaries using an encoder-decoder mechanism. Leveraging a Bidirectional LSTM encoder, it contextualizes each token from the source, while the decoder, supplemented with an attention mechanism, processes the summary. HERMAN checks the quantitative entities' factual consistency against the source using this architecture and re-ranks the summaries to favor those that align with the original text's quantities.
\newline

The task is to reduce the hallucinated entities in LM-based abstractive summarization systems. 
The authors particularly focus on the hallucinations of quantity entities such as datas, numbers, and money amount.
The hallucinated quantity entities that are not supported by the source text, compromise the accuracy and of the generated summaries. 
The challenge is to identify and correct these inaccuracies without compromising the informativeness of the summary.

\begin{figure}[!htbp]
    \centering 
    \includegraphics[width=1.0\linewidth]{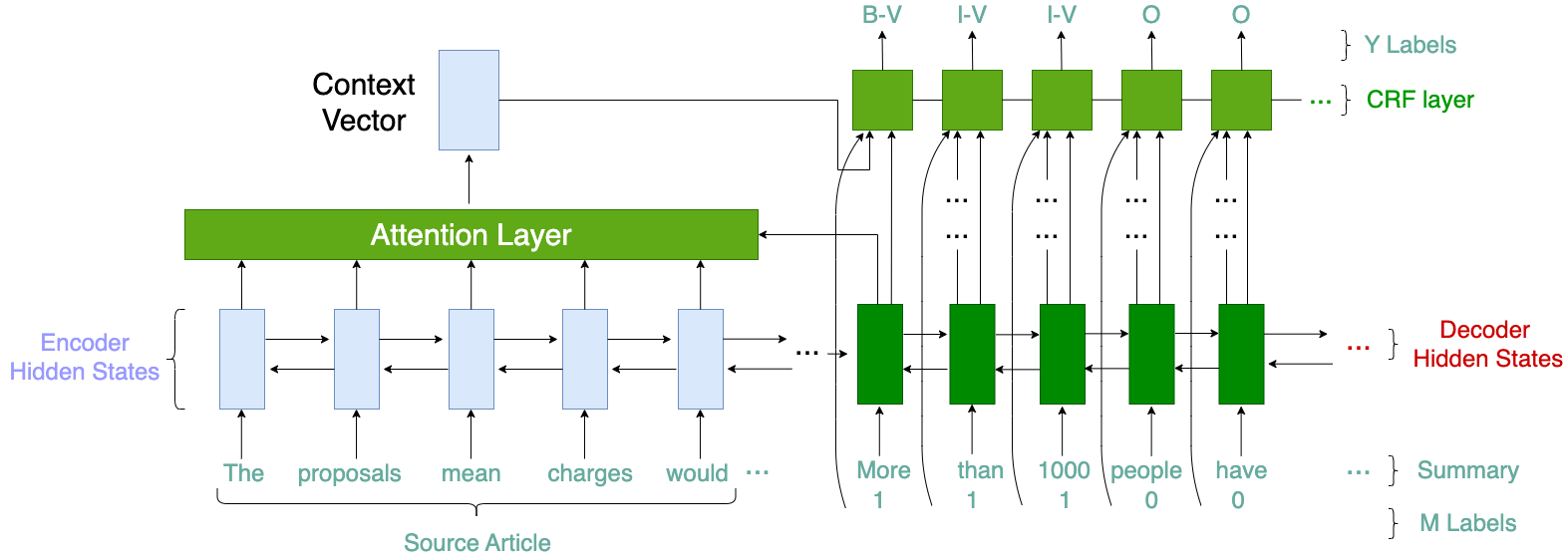} 
    \caption{Architecture of HERMAN. LSTM decoder and Conditional random fields (CRF) modeling methods are employed.~(Figure source:~\cite{zhao2020reducing})} 
    \label{fig:Quantity_0} 
\end{figure}

The authors leveraged the XSum dataset~\cite{narayan-etal-2018-dont}, a well-known summarization dataset that consists of 226,711 news articles accompanied with a one-sentence summary, to produce the dataset for their experiment.
They first filtered the Xsum entries to pinpoint those containing the mentions of quantitative entities such as dates, numbers, and money sums. 
The authors also extracted more instances where these entities were pivotal to the content's meaning. 
The produced dataset contains both filtered set of original data and also synthetic data, where the selected quantitative entities in the summaries are replaced with other entities of the same type from the corresponding source texts. 
This replacement was done randomly, creating variations in the data. 
Each summary was then manually reviewed and annotated with VERIFIED and UNVERIFIED~(with quantitative entities replacement).
The authors proposed HERMAN model to recognize and verify quantity entities.
HERMAN checks the factual consistency of these quantity entities with the source text. 
When coupled with other summarization models, HERMAN re-ranks the generated summaries, giving preference to those that have quantities consistent with the original text. 
The training data for HERMAN contains both original and synthetically created data with the token-level labels (Y labels) and sentence level labels (z labels).
Figure~\ref{fig:Quantity_0} demonstrates the architecture of HERMAN.
The HERMAN model comprises an encoder for the source article and a decoder for the summary.
The encoder produces hidden state representations for every token in the article using a Bidirectional LSTM where each token in the article gets encoded into a contextualized token-level representation. 
The hidden states (context vectors) produced by the encoder are then passed to the attention layer when decoding the summary.
The decoder processes the summary. 
It uses the context vectors from the attention layer and its own hidden states to produce a series of labels (Y labels) for each token in the summary using a combination of a Bidirectional LSTM and a Conditional Random Fields layer, supplemented with an attention mechanism.
The output Y labels classify the quantity entities in the summaries. V: VERIFIED,  U: UNVERIFIED, O:  tokens that are not directly related to any quantity-based entity, B: Begin, I: Inside (BIO tagging scheme). 
Also, the same context vectors are fed into a binary classifier to predict a z label of whether the entire summary is 'VERIFIED' or 'UNVERIFIED'.
%


\begin{table}[t]
\centering
\caption{Results from the automatic evaluation of the XSum test set are organized into three parts, each pertaining to one of the three abstractive summarization systems: BART, BERTSUM, and TCONVS2S. For every system, ROUGE scores are provided for two baseline models, the original model, and two versions of the HERMAN model. Baseline-shortest model chooses the shortest summary. Baseline-max-overlap model selects for the summary with the highest entity overlap with the source document. The term 'avg-Q' represents the average count of quantity entities within each summary.~(Table source:~\cite{zhao2020reducing})} 
\resizebox{0.8\textwidth}{!}{
\begin{tabular}{@{}l|lccc|ccc|ccc|c@{}}
\toprule
 & Model & \multicolumn{1}{l}{R1-R} & \multicolumn{1}{l}{R1-P} & \multicolumn{1}{l|}{R1-F} & \multicolumn{1}{l}{R2-R} & \multicolumn{1}{l}{R2-P} & \multicolumn{1}{l|}{R2-F} & \multicolumn{1}{l}{RL-R} & \multicolumn{1}{l}{RL-P} & \multicolumn{1}{l|}{RL-F} & \multicolumn{1}{l}{avg-Q} \\ \midrule
\multirow{5}{*}{BART} & Baseline-shortest & 45.50 & \textbf{46.95} & \textbf{45.40} & 21.86 & \textbf{22.61} & \textbf{21.83} & 36.80 & \textbf{38.01} & \textbf{36.74} & 0.69 \\
 & Baseline-max-overlap & 49.46 & 41.66 & 44.55 & 23.35 & 19.57 & 20.97 & 39.30 & 33.08 & 35.38 & \textbf{0.95} \\
 & Original & \textbf{49.64} & 41.54 & 44.57 & \textbf{23.43} & 19.50 & 20.96 & \textbf{39.39} & 32.95 & 35.36 & 0.89 \\
 & HERMAN-LOCAL & 48.51 & 42.78 & 44.73 & 22.97 & 20.20 & 21.14 & 38.70 & 34.12 & 35.68 & 0.88 \\
 & HERMAN-GLOBAL & 47.88 & 43.52 & 44.79 & 22.66 & 20.56 & 21.17 & 38.26 & 34.79 & 35.80 & 0.92 \\ \midrule
BERTSUM & Baseline-shortest & 36.78 & \textbf{42.26} & 38.71 & 15.61 & \textbf{17.87} & 16.38 & 29.71 & \textbf{33.91} & 31.16 & 0.62 \\
 & Baseline-max-overlap & 38.17 & 41.25 & 39.01 & \textbf{16.28} & 17.50 & 16.58 & 30.66 & 32.94 & 31.24 & 0.76 \\
 & Original & 38.37 & 40.73 & 38.86 & 16.24 & 17.13 & 16.38 & \textbf{30.75} & 32.44 & 31.04 & 0.65 \\
 & HERMAN-LOCAL & \textbf{38.45} & 40.14 & 38.63 & 16.12 & 16.72 & 16.12 & 30.71 & 31.87 & 30.75 & 0.79 \\
 & HERMAN-GLOBAL & 37.99 & 41.59 & \textbf{39.06} & 16.24 & 17.70 & \textbf{16.65} & 30.59 & 33.28 & \textbf{31.36} & \textbf{0.81} \\ \midrule
\multirow{5}{*}{TCONVS2S} & Baseline-shortest & 27.43 & \textbf{37.28} & 30.99 & 9.84 & \textbf{13.49} & 11.15 & 22.43 & \textbf{30.41} & 25.32 & 0.45 \\
 & Baseline-max-overlap & 30.19 & 34.57 & 31.64 & 10.79 & 12.34 & 11.29 & 24.37 & 27.81 & 25.50 & 0.71 \\
 & Original & \textbf{30.42} & 34.63 & 31.80 & 10.96 & 12.46 & 11.45 & \textbf{24.58} & 27.89 & 25.66 & 0.58 \\
 & HERMAN-LOCAL & 29.95 & 34.50 & 31.43 & 10.59 & 12.16 & 11.09 & 24.17 & 27.72 & 25.31 & 0.75 \\
 & HERMAN-GLOBAL & 30.36 & 34.82 & \textbf{31.85} & \textbf{10.98} & 12.59 & \textbf{11.51} & 24.56 & 28.08 & \textbf{25.72} & \textbf{0.78} \\ \bottomrule
\end{tabular}
}
\label{tab:Quantity_1} 
\end{table}

The HERMAN model has two re-ranking strategies: HERMAN-GLOBAL and HERMAN-LOCAL.
The HERMAN-GLOBAL approach uses the global document-level output label z. 
The HERMAN-LOCAL approach considers the individual token-level y labels to calculate an average verification score for the summary.
Table~\ref{tab:Quantity_1} summarizes the performance of the systems BART~\cite{lewis2020bart}, BERTSUM~\cite{liu-lapata-2019-text}, and TCONVS2S~\cite{narayan-etal-2018-dont} on the XSum~\cite{narayan-etal-2018-dont} test set using ROUGE scores (R1, R2, RL). HERMAN-GLOBAL outperformed other models in ROUGE-1/2/L Precision and F1 scores.

\subsection[Self-contradictory Hallucinations of Large Language Models: Evaluation, Detection
and Mitigation]{Self-contradictory Hallucinations of Large Language Models: Evaluation, Detection
and Mitigation~\cite{mundler2023self}}

\textbf{Github Link: } https://github.com/eth-sri/chatprotect
\newline
\textbf{Task: } Question answering
\newline
\textbf{Core Idea:} The self-contradictory method detects and mitigates hallucinatory content in LLMs by deliberately provoking self-contradictions in generated responses. Using specific input prompts, an LM (gLM) model generates paired statements that are analyzed by another model (aLM) for logical inconsistencies, and once identified, iterative text edits are applied to remove these contradictions.
\newline

The task here is to detect and reduce the produced nonsensical or unfaithful content (the hallucinations) in the responses of large LMs like ChatGPT. 
This method is grounded in the observation that LLMs may produce self-contradictions where they generate two logically inconsistent sentences given the same context. 
By intentionally provoking these inconsistencies, detecting and removing the contradicted content, the authors intend to reduce the hallucinatory content of the LLM-producing responses.
The dataset is the Wikipedia articles of 30 topics such as architecture, artist historical, politics, sports, Television, etc., collected by the authors called the MainTestSet, where the LLM generates 1,000+ sentence pairs.

\begin{figure}[!htbp]
    \centering 
    \includegraphics[width=1.0\linewidth]{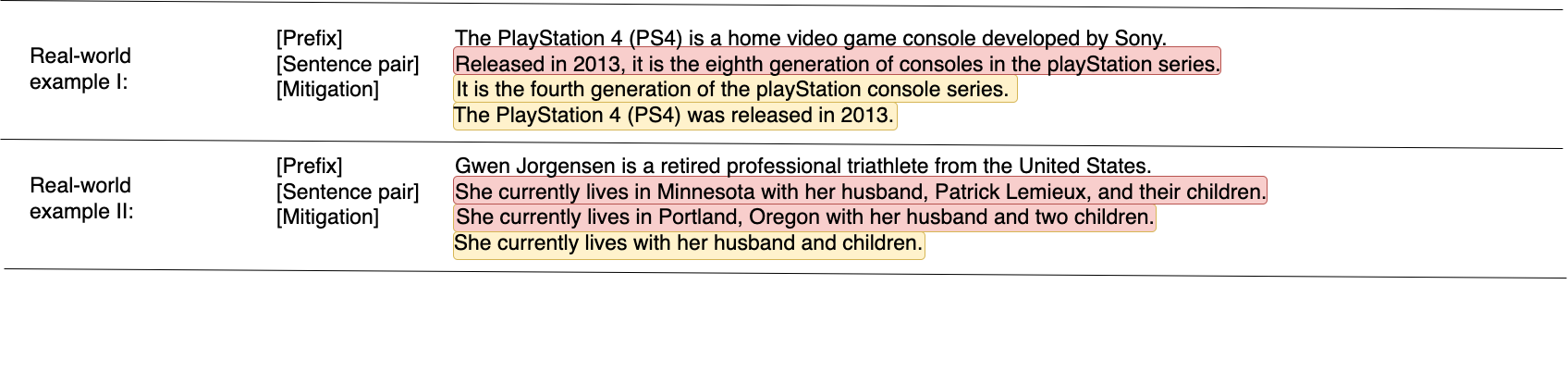} 
    \caption{Two real-world self-contradictory hallucinations produced by ChatGPT based on the provided context are showcased. Sentences highlighted in green are factually correct, while those in red are incorrect. The method proposed identifies and mitigates both hallucinations.~(Figure source:~\cite{mundler2023self})} 
    \label{fig:SC_0} 
\end{figure}

This method is applied to black-box LMs through a system of prompting without needing external knowledge. 
The first step involves intentionally input prompts that are likely to induce contradictory responses to a language model gLM. 
The model gLM will be expected to generate pairs of statements that could potentially be at odds with each other.
The responses of the LM being tested are examined to detect logical inconsistencies or self-contradictions. The presence of contradictions indicates a potential hallucination. 
Following the generation of these pairs, an analyzer model aLM is employed to detect contradictions within the result sentences. 
This analyzer model is trained on recognizing logical inconsistencies to be able to flag the contradiction part of a sentence for further mitigation.
Once contradictions have been detected, the next phase involves the application of iterative text edits. These edits are designed to eliminate or reconcile the identified contradictions while still ensuring the output remains fluent. 
%


\begin{table}[!htbp]
\centering
\caption{Results on the MainTestSet with gMLs, aLMs as control variables. Table source:~\cite{mundler2023self}}
\resizebox{1.0\textwidth}{!}{

\begin{tabular}{@{}lllllll@{}}
\toprule
\multicolumn{7}{l}{Results on the MainTestSet with different gLMs, with ChatGPT as aLM.} \\ \midrule
 & \multicolumn{3}{c}{Detection} & \multicolumn{3}{c}{Mitigation} \\ \cmidrule(l){2-7} 
gLM & P & R & F1 & self-contra. reduced & informative facts retained & perplexity increased \\ \midrule
GPT-4 & 80.1\% & 79.7\% & 79.9\% & 76.3\% & 99.9\% & 0.67 \\
ChatGPT & 84.2\% & 83.2\% & 83.7\% & 89.5\% & 100.8\% & 0.44 \\
Llama2-70B-Chat & 72.9\% & 89.0\% & 80.2\% & 85.4\% & 95.5\% & 0.86 \\
Vicuna-13B & 74.4\% & 83.2\% & 78.6\% & 82.7\% & 95.1\% & 1.78 \\ \midrule
\multicolumn{7}{l}{Results on the MainTestSet with different aLMs. with ChatGPT as gLM.} \\ \midrule
 & \multicolumn{3}{c}{Detection} & \multicolumn{3}{c}{Mitigation} \\ \cmidrule(l){2-7} 
aLM & P & R & F1 & self-contra. reduced & informative facts retained & perplexity increased \\ \midrule
GPT-4 & 91.3\% & 82.1\% & 86.5\% & 79.7\% & 97.7\% & 0.52 \\
ChatGPT & 84.2\% & 83.2\% & 83.7\% & 89.5\% & 100.8\% & 0.44 \\
Llama2-70B-Chat & 95.3\% & 45.8\% & 61.9\% & 44.4\% & 98.8\% & 0.22 \\
Vicuna-13B & 90.0\% & 25.1\% & 39.3\% & 25.5\% & 100.2\% & 0.28 \\ \bottomrule
\end{tabular}
}
\label{tab:Sc_2} 
\end{table}

\begin{figure}[!htbp]
    \centering 
    \includegraphics[width=1.0\linewidth]{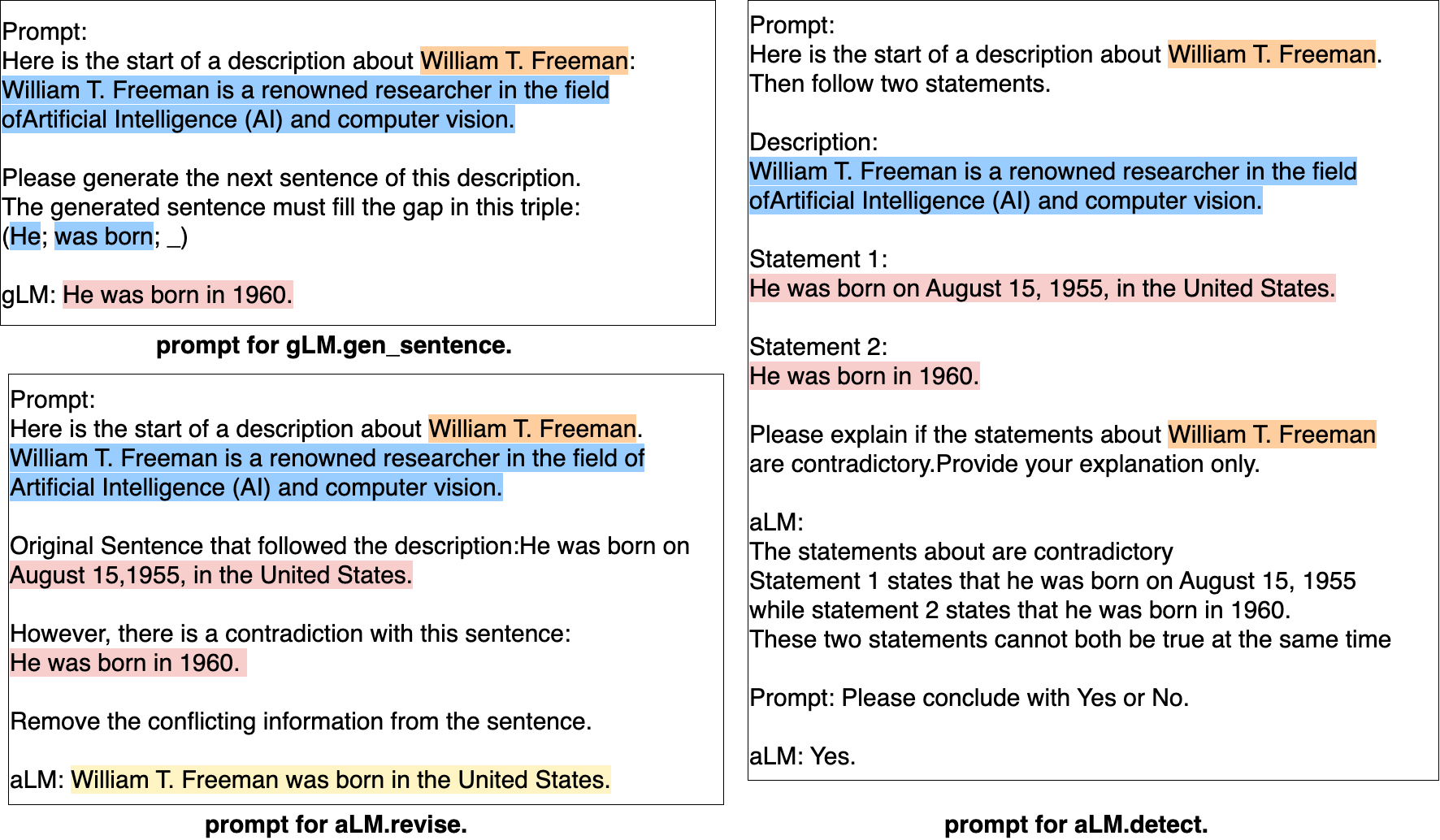} 
    \caption{The prompts for the entire process.~(Figure source:~\cite{mundler2023self})} 
    \label{fig:Sc_1} 
\end{figure}

Both the gLM and aLM can be any of the state-of-the-art language models. 
In the experiment, ChatGPT (3.5), GPT-4, Llama2-70B-Chat, and Vicuna-13B are utilized and compared.
%
%
Table~\ref{tab:Sc_2} reveals that when using ChatGPT as aLM,
the proposed approach is consistently effective across different gLMs.
The detection reaches a F1 score of \textasciitilde{}80\%. The mitigation can remove a self-contradictions up to 89.5\% with ChatGPT as gLM. 
The approach also maintains informativeness and ends up with small text fluency harming (small perplexity increase).
When using ChatGPT as gLM and comparing the language models as aLM, Vicuna-13B performed notably poorly as an aLM in terms of detection metrics. 
Both two open-source language model Llama2-70B-Chat and Vicuna-13B shows low ability to reduce the self-contradictions.
Both proprietary LMs GPT-4 and ChatGPT demonstrated strong performance in terms of detection and removing self-contradictions. 
%
%

\section{Conclusion}
In this report, we briefly review state-of-the-art hallucination detection and mitigation methods and report some of the reproduced results of these papers. For hallucination detection, the reviewed papers mainly come from two categories: token-level detection and sentence-level detection, with each type of methods focusing on different granularity of the detection. 
The mitigation approaches often concern with either mitigate the hallucinations pre-generation or post-generation. Pre-generation methods typically involves tricks in training/finetuning while post-generation approaches often consider leveraging knowledge retrieval methods. In addition to reviewing the literature, we have also reproduced some of the results from these papers. Most of the cases we have enough data/hyper-parameter settings to reproduce the results almost perfectly. However, there are cases (such as Hades), where the lack the hyper-parameter information brings some difficulties in reproducing the results exactly.  
We hope that this report can be a valuable reference for future hallucination detection and mitigation works, especially in real-world applications. 
\newpage
\bibliographystyle{plain} 
\bibliography{reference}

\begin{thebibliography}{10}

\bibitem{banerjee-lavie-2005-meteor}
Satanjeev Banerjee and Alon Lavie.
\newblock {METEOR}: An automatic metric for {MT} evaluation with improved
  correlation with human judgments.
\newblock In {\em Proceedings of the {ACL} Workshop on Intrinsic and Extrinsic
  Evaluation Measures for Machine Translation and/or Summarization}, pages
  65--72, Ann Arbor, Michigan, June 2005. Association for Computational
  Linguistics.

\bibitem{bordes2013translating}
Antoine Bordes, Nicolas Usunier, Alberto Garcia-Duran, Jason Weston, and Oksana
  Yakhnenko.
\newblock Translating embeddings for modeling multi-relational data.
\newblock {\em Advances in neural information processing systems}, 26, 2013.

\bibitem{bowman2016generating}
Samuel Bowman, Luke Vilnis, Oriol Vinyals, Andrew Dai, Rafal Jozefowicz, and
  Samy Bengio.
\newblock Generating sentences from a continuous space.
\newblock In {\em Proceedings of the 20th SIGNLL Conference on Computational
  Natural Language Learning}, pages 10--21, 2016.

\bibitem{brown2020language}
Tom Brown, Benjamin Mann, Nick Ryder, Melanie Subbiah, Jared~D Kaplan, Prafulla
  Dhariwal, Arvind Neelakantan, Pranav Shyam, Girish Sastry, Amanda Askell,
  et~al.
\newblock Language models are few-shot learners.
\newblock {\em Advances in neural information processing systems},
  33:1877--1901, 2020.

\bibitem{cao-etal-2022-hallucinated}
Meng Cao, Yue Dong, and Jackie Cheung.
\newblock Hallucinated but factual! inspecting the factuality of hallucinations
  in abstractive summarization.
\newblock In {\em Proceedings of the 60th Annual Meeting of the Association for
  Computational Linguistics (Volume 1: Long Papers)}, pages 3340--3354, Dublin,
  Ireland, May 2022. Association for Computational Linguistics.

\bibitem{conneau2020unsupervised}
Alexis Conneau, Kartikay Khandelwal, Naman Goyal, Vishrav Chaudhary, Guillaume
  Wenzek, Francisco Guzm{\'a}n, {\'E}douard Grave, Myle Ott, Luke Zettlemoyer,
  and Veselin Stoyanov.
\newblock Unsupervised cross-lingual representation learning at scale.
\newblock In {\em Proceedings of the 58th Annual Meeting of the Association for
  Computational Linguistics}, pages 8440--8451, 2020.

\bibitem{dinan2018wizard}
Emily Dinan, Stephen Roller, Kurt Shuster, Angela Fan, Michael Auli, and Jason
  Weston.
\newblock Wizard of wikipedia: Knowledge-powered conversational agents.
\newblock In {\em International Conference on Learning Representations}, 2018.

\bibitem{durmus-etal-2020-feqa}
Esin Durmus, He~He, and Mona Diab.
\newblock {FEQA}: A question answering evaluation framework for faithfulness
  assessment in abstractive summarization.
\newblock In {\em Proceedings of the 58th Annual Meeting of the Association for
  Computational Linguistics}, pages 5055--5070, Online, July 2020. Association
  for Computational Linguistics.

\bibitem{dziri2021neural}
Nouha Dziri, Andrea Madotto, Osmar~R Zaiane, and Avishek~Joey Bose.
\newblock Neural path hunter: Reducing hallucination in dialogue systems via
  path grounding.
\newblock In {\em Proceedings of the 2021 Conference on Empirical Methods in
  Natural Language Processing}, pages 2197--2214, 2021.

\bibitem{fabbri2020summeval}
Alexander~R Fabbri, Wojciech Kry{\'s}ci{\'n}ski, Bryan McCann, Caiming Xiong,
  Richard Socher, and Dragomir Radev.
\newblock Summeval: Re-evaluating summarization evaluation.
\newblock {\em arXiv preprint arXiv:2007.12626}, 2020.

\bibitem{goyal-durrett-2020-evaluating}
Tanya Goyal and Greg Durrett.
\newblock Evaluating factuality in generation with dependency-level entailment.
\newblock In {\em Findings of the Association for Computational Linguistics:
  EMNLP 2020}, pages 3592--3603, Online, November 2020. Association for
  Computational Linguistics.

\bibitem{gu2020generating}
Xiaotao Gu, Yuning Mao, Jiawei Han, Jialu Liu, You Wu, Cong Yu, Daniel Finnie,
  Hongkun Yu, Jiaqi Zhai, and Nicholas Zukoski.
\newblock Generating representative headlines for news stories.
\newblock In {\em 29th International World Wide Web Conference, WWW 2020},
  pages 1773--1784. Association for Computing Machinery, Inc, 2020.

\bibitem{guo2020wiki}
Mandy Guo, Zihang Dai, Denny Vrande{\v{c}}i{\'c}, and Rami Al-Rfou.
\newblock Wiki-40b: Multilingual language model dataset.
\newblock In {\em Proceedings of the Twelfth Language Resources and Evaluation
  Conference}, pages 2440--2452, 2020.

\bibitem{he2022debertav3}
Pengcheng He, Jianfeng Gao, and Weizhu Chen.
\newblock Debertav3: Improving deberta using electra-style pre-training with
  gradient-disentangled embedding sharing.
\newblock In {\em The Eleventh International Conference on Learning
  Representations}, 2022.

\bibitem{spacy2}
Matthew Honnibal and Ines Montani.
\newblock {spaCy 2}: Natural language understanding with {B}loom embeddings,
  convolutional neural networks and incremental parsing.
\newblock To appear, 2017.

\bibitem{humeau2019poly}
Samuel Humeau, Kurt Shuster, Marie-Anne Lachaux, and Jason Weston.
\newblock Poly-encoders: Architectures and pre-training strategies for fast and
  accurate multi-sentence scoring.
\newblock In {\em International Conference on Learning Representations}, 2019.

\bibitem{izacard2021leveraging}
Gautier Izacard and {\'E}douard Grave.
\newblock Leveraging passage retrieval with generative models for open domain
  question answering.
\newblock In {\em Proceedings of the 16th Conference of the European Chapter of
  the Association for Computational Linguistics: Main Volume}, pages 874--880,
  2021.

\bibitem{ji2023rho}
Ziwei Ji, Zihan Liu, Nayeon Lee, Tiezheng Yu, Bryan Wilie, Min Zeng, and
  Pascale Fung.
\newblock Rho: Reducing hallucination in open-domain dialogues with knowledge
  grounding.
\newblock In {\em Findings of the Association for Computational Linguistics:
  ACL 2023}, pages 4504--4522, 2023.

\bibitem{karpukhin2020dense}
Vladimir Karpukhin, Barlas Oguz, Sewon Min, Patrick Lewis, Ledell Wu, Sergey
  Edunov, Danqi Chen, and Wen-tau Yih.
\newblock Dense passage retrieval for open-domain question answering.
\newblock In {\em Proceedings of the 2020 Conference on Empirical Methods in
  Natural Language Processing (EMNLP)}, pages 6769--6781, 2020.

\bibitem{kenton2019bert}
Jacob Devlin Ming-Wei~Chang Kenton and Lee~Kristina Toutanova.
\newblock Bert: Pre-training of deep bidirectional transformers for language
  understanding.
\newblock In {\em Proceedings of NAACL-HLT}, pages 4171--4186, 2019.

\bibitem{laban-etal-2022-summac}
Philippe Laban, Tobias Schnabel, Paul~N. Bennett, and Marti~A. Hearst.
\newblock {S}umma{C}: Re-visiting {NLI}-based models for inconsistency
  detection in summarization.
\newblock {\em Transactions of the Association for Computational Linguistics},
  10:163--177, 2022.

\bibitem{lebret-etal-2016-neural}
R{\'e}mi Lebret, David Grangier, and Michael Auli.
\newblock Neural text generation from structured data with application to the
  biography domain.
\newblock In {\em Proceedings of the 2016 Conference on Empirical Methods in
  Natural Language Processing}, pages 1203--1213, Austin, Texas, November 2016.
  Association for Computational Linguistics.

\bibitem{lewis2020bart}
Mike Lewis, Yinhan Liu, Naman Goyal, Marjan Ghazvininejad, Abdelrahman Mohamed,
  Omer Levy, Veselin Stoyanov, and Luke Zettlemoyer.
\newblock Bart: Denoising sequence-to-sequence pre-training for natural
  language generation, translation, and comprehension.
\newblock In {\em Proceedings of the 58th Annual Meeting of the Association for
  Computational Linguistics}, pages 7871--7880, 2020.

\bibitem{HaluEval}
Junyi Li, Xiaoxue Cheng, Wayne~Xin Zhao, Jian-Yun Nie, and Ji-Rong Wen.
\newblock Halueval: A large-scale hallucination evaluation benchmark for large
  language models, 2023.

\bibitem{lin-2004-rouge}
Chin-Yew Lin.
\newblock {ROUGE}: A package for automatic evaluation of summaries.
\newblock In {\em Text Summarization Branches Out}, pages 74--81, Barcelona,
  Spain, July 2004. Association for Computational Linguistics.

\bibitem{liu2022token}
Tianyu Liu, Yizhe Zhang, Chris Brockett, Yi~Mao, Zhifang Sui, Weizhu Chen, and
  William~B Dolan.
\newblock A token-level reference-free hallucination detection benchmark for
  free-form text generation.
\newblock In {\em Proceedings of the 60th Annual Meeting of the Association for
  Computational Linguistics (Volume 1: Long Papers)}, pages 6723--6737, 2022.

\bibitem{liu-lapata-2019-text}
Yang Liu and Mirella Lapata.
\newblock Text summarization with pretrained encoders.
\newblock In Kentaro Inui, Jing Jiang, Vincent Ng, and Xiaojun Wan, editors,
  {\em Proceedings of the 2019 Conference on Empirical Methods in Natural
  Language Processing and the 9th International Joint Conference on Natural
  Language Processing (EMNLP-IJCNLP)}, pages 3730--3740, Hong Kong, China,
  November 2019. Association for Computational Linguistics.

\bibitem{manakul2023selfcheckgpt}
Potsawee Manakul, Adian Liusie, and Mark J.~F. Gales.
\newblock Selfcheckgpt: Zero-resource black-box hallucination detection for
  generative large language models, 2023.

\bibitem{manakul2023mqag}
Potsawee Manakul, Adian Liusie, and Mark~JF Gales.
\newblock Mqag: Multiple-choice question answering and generation for assessing
  information consistency in summarization.
\newblock {\em arXiv preprint arXiv:2301.12307}, 2023.

\bibitem{maynez2020faithfulness}
Joshua Maynez, Shashi Narayan, Bernd Bohnet, and Ryan McDonald.
\newblock On faithfulness and factuality in abstractive summarization.
\newblock In {\em Proceedings of the 58th Annual Meeting of the Association for
  Computational Linguistics}, pages 1906--1919, 2020.

\bibitem{miao2021prevent}
Mengqi Miao, Fandong Meng, Yijin Liu, Xiao-Hua Zhou, and Jie Zhou.
\newblock Prevent the language model from being overconfident in neural machine
  translation.
\newblock In {\em Proceedings of the 59th Annual Meeting of the Association for
  Computational Linguistics and the 11th International Joint Conference on
  Natural Language Processing (Volume 1: Long Papers)}, pages 3456--3468, 2021.

\bibitem{moon2019opendialkg}
Seungwhan Moon, Pararth Shah, Anuj Kumar, and Rajen Subba.
\newblock Opendialkg: Explainable conversational reasoning with attention-based
  walks over knowledge graphs.
\newblock In {\em Proceedings of the 57th annual meeting of the association for
  computational linguistics}, pages 845--854, 2019.

\bibitem{mundler2023self}
Niels M{\"u}ndler, Jingxuan He, Slobodan Jenko, and Martin Vechev.
\newblock Self-contradictory hallucinations of large language models:
  Evaluation, detection and mitigation.
\newblock {\em arXiv preprint arXiv:2305.15852}, 2023.

\bibitem{nallapati2016abstractive}
Ramesh Nallapati, Bowen Zhou, Cicero dos Santos, Caglar Gulcehre, and Bing
  Xiang.
\newblock Abstractive text summarization using sequence-to-sequence rnns and
  beyond.
\newblock In {\em Proceedings of The 20th SIGNLL Conference on Computational
  Natural Language Learning}, page 280. Association for Computational
  Linguistics, 2016.

\bibitem{narayan-etal-2018-dont}
Shashi Narayan, Shay~B. Cohen, and Mirella Lapata.
\newblock Don{'}t give me the details, just the summary! topic-aware
  convolutional neural networks for extreme summarization.
\newblock In {\em Proceedings of the 2018 Conference on Empirical Methods in
  Natural Language Processing}, pages 1797--1807, Brussels, Belgium,
  October-November 2018. Association for Computational Linguistics.

\bibitem{pagnoni-etal-2021-understanding}
Artidoro Pagnoni, Vidhisha Balachandran, and Yulia Tsvetkov.
\newblock Understanding factuality in abstractive summarization with {FRANK}: A
  benchmark for factuality metrics.
\newblock In Kristina Toutanova, Anna Rumshisky, Luke Zettlemoyer, Dilek
  Hakkani-Tur, Iz~Beltagy, Steven Bethard, Ryan Cotterell, Tanmoy Chakraborty,
  and Yichao Zhou, editors, {\em Proceedings of the 2021 Conference of the
  North American Chapter of the Association for Computational Linguistics:
  Human Language Technologies}, pages 4812--4829, Online, June 2021.
  Association for Computational Linguistics.

\bibitem{papineni2002bleu}
Kishore Papineni, Salim Roukos, Todd Ward, and Wei-Jing Zhu.
\newblock Bleu: a method for automatic evaluation of machine translation.
\newblock In {\em Proceedings of the 40th annual meeting of the Association for
  Computational Linguistics}, pages 311--318, 2002.

\bibitem{post-2018-call}
Matt Post.
\newblock A call for clarity in reporting {BLEU} scores.
\newblock In {\em Proceedings of the Third Conference on Machine Translation:
  Research Papers}, pages 186--191, Belgium, Brussels, October 2018.
  Association for Computational Linguistics.

\bibitem{pu2021learning}
Amy Pu, Hyung~Won Chung, Ankur~P Parikh, Sebastian Gehrmann, and Thibault
  Sellam.
\newblock Learning compact metrics for mt.
\newblock In {\em Proceedings of EMNLP}, 2021.

\bibitem{radford2019language}
Alec Radford, Jeff Wu, Rewon Child, David Luan, Dario Amodei, and Ilya
  Sutskever.
\newblock Language models are unsupervised multitask learners.
\newblock https://api.semanticscholar.org/CorpusID:160025533, 2019.

\bibitem{rashkin2021increasing}
Hannah Rashkin, David Reitter, Gaurav~Singh Tomar, and Dipanjan Das.
\newblock Increasing faithfulness in knowledge-grounded dialogue with
  controllable features.
\newblock In {\em Proceedings of the 59th Annual Meeting of the Association for
  Computational Linguistics and the 11th International Joint Conference on
  Natural Language Processing (Volume 1: Long Papers)}, pages 704--718, 2021.

\bibitem{scialom-etal-2021-questeval}
Thomas Scialom, Paul-Alexis Dray, Sylvain Lamprier, Benjamin Piwowarski, Jacopo
  Staiano, Alex Wang, and Patrick Gallinari.
\newblock {Q}uest{E}val: Summarization asks for fact-based evaluation.
\newblock In {\em Proceedings of the 2021 Conference on Empirical Methods in
  Natural Language Processing}, pages 6594--6604, Online and Punta Cana,
  Dominican Republic, November 2021. Association for Computational Linguistics.

\bibitem{sellam2020bleurt}
Thibault Sellam, Dipanjan Das, and Ankur~P Parikh.
\newblock Bleurt: Learning robust metrics for text generation.
\newblock In {\em Proceedings of ACL}, 2020.

\bibitem{shen2023misleading}
Jiaming Shen, Jialu Liu, Dan Finnie, Negar Rahmati, Michael Bendersky, and Marc
  Najork.
\newblock " why is this misleading?": Detecting news headline hallucinations
  with explanations.
\newblock {\em arXiv preprint arXiv:2302.05852}, 2023.

\bibitem{shuster2021retrieval}
Kurt Shuster, Spencer Poff, Moya Chen, Douwe Kiela, and Jason Weston.
\newblock Retrieval augmentation reduces hallucination in conversation.
\newblock In {\em Findings of the Association for Computational Linguistics:
  EMNLP 2021}, pages 3784--3803, 2021.

\bibitem{son-etal-2022-harim}
Seonil~(Simon) Son, Junsoo Park, Jeong-in Hwang, Junghwa Lee, Hyungjong Noh,
  and Yeonsoo Lee.
\newblock {H}a{R}i{M}$^+$: Evaluating summary quality with hallucination risk.
\newblock In {\em Proceedings of the 2nd Conference of the Asia-Pacific Chapter
  of the Association for Computational Linguistics and the 12th International
  Joint Conference on Natural Language Processing}, pages 895--924, Online
  only, November 2022. Association for Computational Linguistics.

\bibitem{sun2023contrastive}
Weiwei Sun, Zhengliang Shi, Shen Gao, Pengjie Ren, Maarten de~Rijke, and
  Zhaochun Ren.
\newblock Contrastive learning reduces hallucination in conversations.
\newblock In {\em Proceedings of the AAAI Conference on Artificial
  Intelligence}, volume~37, pages 13618--13626, 2023.

\bibitem{alpaca}
Rohan Taori, Ishaan Gulrajani, Tianyi Zhang, Yann Dubois, Xuechen Li, Carlos
  Guestrin, Percy Liang, and Tatsunori~B. Hashimoto.
\newblock Stanford alpaca: An instruction-following llama model.
\newblock \url{https://github.com/tatsu-lab/stanford_alpaca}, 2023.

\bibitem{touvron2023llama}
Hugo Touvron, Thibaut Lavril, Gautier Izacard, Xavier Martinet, Marie-Anne
  Lachaux, Timoth{\'e}e Lacroix, Baptiste Rozi{\`e}re, Naman Goyal, Eric
  Hambro, Faisal Azhar, et~al.
\newblock Llama: Open and efficient foundation language models.
\newblock {\em arXiv preprint arXiv:2302.13971}, 2023.

\bibitem{wang-etal-2020-asking}
Alex Wang, Kyunghyun Cho, and Mike Lewis.
\newblock Asking and answering questions to evaluate the factual consistency of
  summaries.
\newblock In Dan Jurafsky, Joyce Chai, Natalie Schluter, and Joel Tetreault,
  editors, {\em Proceedings of the 58th Annual Meeting of the Association for
  Computational Linguistics}, pages 5008--5020, Online, July 2020. Association
  for Computational Linguistics.

\bibitem{wang2020go}
Yong Wang, Longyue Wang, Shuming Shi, Victor~OK Li, and Zhaopeng Tu.
\newblock Go from the general to the particular: Multi-domain translation with
  domain transformation networks.
\newblock In {\em Proceedings of the AAAI Conference on Artificial
  Intelligence}, volume~34, pages 9233--9241, 2020.

\bibitem{wang2014knowledge}
Zhen Wang, Jianwen Zhang, Jianlin Feng, and Zheng Chen.
\newblock Knowledge graph embedding by translating on hyperplanes.
\newblock In {\em Proceedings of the AAAI conference on artificial
  intelligence}, volume~28, 2014.

\bibitem{wei2022chain}
Jason Wei, Xuezhi Wang, Dale Schuurmans, Maarten Bosma, Fei Xia, Ed~Chi, Quoc~V
  Le, Denny Zhou, et~al.
\newblock Chain-of-thought prompting elicits reasoning in large language
  models.
\newblock {\em Advances in Neural Information Processing Systems},
  35:24824--24837, 2022.

\bibitem{yang2019xlnet}
Zhilin Yang, Zihang Dai, Yiming Yang, Jaime Carbonell, Russ~R Salakhutdinov,
  and Quoc~V Le.
\newblock Xlnet: Generalized autoregressive pretraining for language
  understanding.
\newblock {\em Advances in neural information processing systems}, 32, 2019.

\bibitem{NEURIPS2021_bartscore}
Weizhe Yuan, Graham Neubig, and Pengfei Liu.
\newblock Bartscore: Evaluating generated text as text generation.
\newblock In M.~Ranzato, A.~Beygelzimer, Y.~Dauphin, P.S. Liang, and J.~Wortman
  Vaughan, editors, {\em Advances in Neural Information Processing Systems},
  volume~34, pages 27263--27277. Curran Associates, Inc., 2021.

\bibitem{zha-etal-2023-alignscore}
Yuheng Zha, Yichi Yang, Ruichen Li, and Zhiting Hu.
\newblock {A}lign{S}core: Evaluating factual consistency with a unified
  alignment function.
\newblock In {\em Proceedings of the 61st Annual Meeting of the Association for
  Computational Linguistics (Volume 1: Long Papers)}, pages 11328--11348,
  Toronto, Canada, July 2023. Association for Computational Linguistics.

\bibitem{bert-score}
Tianyi Zhang*, Varsha Kishore*, Felix Wu*, Kilian~Q. Weinberger, and Yoav
  Artzi.
\newblock Bertscore: Evaluating text generation with bert.
\newblock In {\em International Conference on Learning Representations}, 2020.

\bibitem{zhao2020reducing}
Zheng Zhao, Shay~B Cohen, and Bonnie Webber.
\newblock Reducing quantity hallucinations in abstractive summarization.
\newblock In {\em Findings of the Association for Computational Linguistics:
  EMNLP 2020}, pages 2237--2249, 2020.

\bibitem{zhou2021detecting}
Chunting Zhou, Graham Neubig, Jiatao Gu, Mona Diab, Francisco Guzm{\'a}n, Luke
  Zettlemoyer, and Marjan Ghazvininejad.
\newblock Detecting hallucinated content in conditional neural sequence
  generation.
\newblock In {\em Findings of the Association for Computational Linguistics:
  ACL-IJCNLP 2021}, pages 1393--1404, 2021.

\bibitem{zhuang-etal-2021-robustly}
Liu Zhuang, Lin Wayne, Shi Ya, and Zhao Jun.
\newblock A robustly optimized {BERT} pre-training approach with post-training.
\newblock In {\em Proceedings of the 20th Chinese National Conference on
  Computational Linguistics}, pages 1218--1227, Huhhot, China, August 2021.
  Chinese Information Processing Society of China.

\end{thebibliography}

\end{document}